\documentclass[journal]{IEEEtran}

\usepackage[english]{babel}
\usepackage[utf8x]{inputenc}
\usepackage[T1]{fontenc}

\usepackage{amsmath,amssymb,mathrsfs,bbm}
\usepackage{graphicx}
\usepackage[colorinlistoftodos]{todonotes}
\usepackage[colorlinks=true, allcolors=blue]{hyperref}
\usepackage{multirow}
\usepackage[lined,linesnumbered,ruled]{algorithm2e}

\usepackage{psfrag,epsfig,graphics}
\usepackage{amsmath,amsthm,amssymb,multirow}
\usepackage{mathbbol}
\usepackage{amssymb}

\DeclareSymbolFontAlphabet{\amsmathbb}{AMSb}%

\newcommand{\cp}[1]{\ifmmode {\mathcal{#1}}\else ${\mathcal{#1}}$\fi}
	
\newcommand{\bA}{\boldsymbol{A}}
\newcommand{\bB}{\boldsymbol{B}}

\newcommand{\bD}{\boldsymbol{D}}

\newcommand{\bF}{\boldsymbol{F}}

\newcommand{\bH}{\boldsymbol{H}}
\newcommand{\bI}{\boldsymbol{I}}

\newcommand{\bM}{\boldsymbol{M}}
\newcommand{\bN}{\boldsymbol{N}}
\newcommand{\bP}{\boldsymbol{P}}

\newcommand{\bR}{\boldsymbol{R}}

\newcommand{\bU}{\boldsymbol{U}}

\newcommand{\bX}{\boldsymbol{X}}
\newcommand{\bY}{\boldsymbol{Y}}
\newcommand{\bZ}{\boldsymbol{Z}}

\newcommand{\ba}{\boldsymbol{a}}
\newcommand{\bb}{\boldsymbol{b}}
\newcommand{\bc}{\boldsymbol{c}}

\newcommand{\bh}{\boldsymbol{h}}

\newcommand{\by}{\boldsymbol{y}}

\newcommand{\bu}{\boldsymbol{u}}

\newcommand{\bx}{\boldsymbol{x}}
\newcommand{\bw}{\boldsymbol{w}}

\newcommand{\bZhat}{\,\widehat{\!\bZ}}

\newcommand{\calF}{\mathcal{F}}

\newcommand{\bbA}{\mathbb{A}}
\newcommand{\bbB}{\mathbb{B}}

\newcommand{\bbR}{\mathbb{R}}

\newcommand{\bbU}{\mathbb{U}}

\newcommand{\bOmega}{\boldsymbol{\Omega}}
\newcommand{\bPsi}{\boldsymbol{\Psi}}

\newcommand{\bTheta}{\boldsymbol{\Theta}}

\newcommand{\cb}[1]{\boldsymbol{#1}}

\newcommand{\Ex}{\mathbb{E}}

\usepackage{color}  

\definecolor{darkgreen}{rgb}{0., 0.4, 0.}

\title{Super-Resolution for Hyperspectral and Multispectral Image Fusion Accounting for Seasonal Spectral Variability}

\author{Ricardo~Augusto~Borsoi, Tales Imbiriba,~\IEEEmembership{Member,~IEEE,} Jos\'e~Carlos~Moreira~Bermudez,~\IEEEmembership{Senior~Member,~IEEE}
\thanks{This work has been supported by the National Council for Scientific and Technological Development (CNPq) under grants 304250/2017-1, 409044/2018-0, 141271/2017-5 and 204991/2018-8, and by the Brazilian Education Ministry (CAPES) under grant PNPD/1811213.}
\thanks{R.A. Borsoi is with the Department of Electrical Engineering, Federal University of Santa Catarina (DEE--UFSC), Florian\'opolis, SC, Brazil, and with the Lagrange Laboratory, Universit\'e  C\^ote  d'Azur, Nice, France. e-mail: \mbox{raborsoi@gmail.com}.
T. Imbiriba was with the DEE--UFSC, Florian\'opolis, SC, Brazil, and is with the ECE department of the Northeastern University, Boston, MA, USA. e-mail: \mbox{talesim@ece.neu.edu}. 
J.C.M. Bermudez is with the DEE--UFSC, Florian\'opolis, SC, Brazil, and with the Graduate Program on Electronic Engineering and Computing, Catholic University of Pelotas (UCPel) Pelotas, Brazil. e-mail: \mbox{j.bermudez@ieee.org}.}
\thanks{Manuscript received Month day, year; revised Month day, year.}}

\begin{document}
\maketitle

\begin{abstract}
Image fusion combines data from different heterogeneous sources to obtain more precise information about an underlying scene. Hyperspectral-multispectral (HS-MS) image fusion is currently attracting great interest in remote sensing since it allows the generation of high spatial resolution HS images, circumventing the main limitation of this imaging modality. Existing HS-MS fusion algorithms, however, neglect the spectral variability often existing between images acquired at different time instants. This time difference causes variations in spectral signatures of the underlying constituent materials due to different acquisition and seasonal conditions. This paper introduces a novel HS-MS image fusion strategy that combines an unmixing-based formulation with an explicit parametric model for typical spectral variability between the two images. Simulations with synthetic and real data show that the proposed strategy leads to a significant performance improvement under spectral variability and state-of-the-art performance otherwise.
\end{abstract}


\section{Introduction}


Hyperspectral (HS) imaging devices are non-conventional sensors that sample the observed electromagnetic spectra at hundreds of contiguous wavelength intervals. 
This spectral resolution makes HS imaging an important tool for identification of the materials present in a scene, what has attracted interest in the past two decades. Hyperspectral images (HSIs) are now at the core of a vast and increasing number of remote sensing applications such as land use analysis, mineral detection, environment monitoring, and field surveillance~\mbox{\cite{Bioucas-Dias-2013-ID307,Manolakis:2002p5224}}.

However, the high spectral resolution of HSIs does not come without a compromise. Since the radiated light observed at the sensor must be divided into a large number of spectral bands, the size of each HSI pixel must be large enough to attain a minimum signal to noise ratio. This leads to images with low spatial resolution~\cite{shaw2003reviewSpecImaging}.
Multispectral (MS) sensors, on the other hand, provide images with much higher spatial resolution, albeit with a small number of spectral bands.

One approach to obtain images with high spatial and spectral resolution consists in combining HS and MS images (MSI) of the same scene, resulting in the so-called HS-MS image fusion problem~\cite{yokoya2017HS_MS_fusinoRev}. 
Several algorithms have been proposed to solve this problem. Early approaches were based on component substitution or on multiresolution analysis. In the former, a component of the HS image in some feature space is replaced by a corresponding component of the MS image~\cite{aiazzi2007componentSubstitutionHSPan}. In the latter, high-frequency spatial details obtained from the MS image are injected into the HS image~\cite{liu2000MRA_HS_MS_fusion,aiazzi2006GLP_HS}.
These techniques generalize Pansharpening algorithms, which combine an HSI with a single complementary monochromatic image~\cite{loncan2015pansharpeningReview}.

More recently, subspace-based formulations have become the leading approach to solve this problem, exploring the natural low-dimensional representation of HSIs as a linear combination of a small set of basis vectors or spectral signatures~\cite{yokoya2017HS_MS_fusinoRev,Keshava:2002p5667,imbiriba2018_ULTRA,Borsoi2017_multiscale}.
Many approaches have been proposed to perform image fusion under this formulation.
For instance, Bayesian formulations have been proposed to solve this problem as a maximum a posteriori estimation problem~\cite{hardie2004hs_ms_fusionMAP}, as the solution to a Sylvester equation~\cite{wei2015hs_ms_fusionSylvesterEq}, or by considering different forms of sparse representations on learned dictionaries~\cite{wei2015hs_ms_fusionBayesianSparse,akhtar2015hs_ms_fusionBayesianSparse}.
Other works propose a matrix factorization formulation, using for instance sparse~\cite{kawakami2011hs_ms_fusionMFsparse,wycoff2013hs_ms_fusionMFsparse} or spatial regularizations~\cite{simoes2015HySure}, or processing local image patches individually~\cite{veganzones2016hs_ms_fusionMFlocalLowRank}.
Other approaches use unsupervised formulations which jointly estimate the set of basis vectors and their coefficients~\cite{yokoya2012coupledNMF,lanaras2015hs_ms_fusionMFprojectedGrad}.
More recently, tensor factorization methods have also been proposed to solve the image fusion problem by exploring the natural representation of HSIs as 3-dimensional tensors~\cite{li2018hs_ms_fusionTensorFactorization,kanatsoulis2018hs_ms_fusionTensorFactorization}.

Although platforms carrying both HS and MS imaging sensors are becoming more common in recent years, their number is still considerably limited~\cite{eckardt2015DESIS_satelliteSpecs,kaufmann2006EnMAP_satelliteSpecs}.
However, the increasing number of optical satellites orbiting the Earth (e.g. Sentinel, Orbview, Landsat and Quickbird missions) provides a large amount of MS data, which can be exploited to perform image fusion using HS images acquired on board of different missions~\cite{yokoya2017HS_MS_fusinoRev}. This scenario provides the greatest potential for the use of HS-MS fusion in practical applications.
Furthermore, the short revisit cycles of these satellites provide MS images at short subsequent time instants. These MS images  can be used to generate HS image sequences with a high temporal resolution (what has been already done with Landsat and MODIS data~\cite{gao2006fusion_MODIS_denseTimeSeries_exp,roy2008fusion_MODIS_denseTimeSeries_exp,hilker2009fusion_MODIS_denseTimeSeries_exp,hilker2009fusion_MODIS_denseTimeSeries_exp2,emelyanova2013fusion_MODIS_denseTimeSeries_exp}).

Though combining images from different sensors provides a great opportunity for applying image fusion algorithms, it introduces a challenge that has been largely neglected. Since the HS and MS images are observed at different time instants, their acquisition conditions (e.g. atmospheric and illumination) can be significantly different from one another. Furthermore, seasonal changes can make the spectral signature of the materials in the HS image change significantly from those in the corresponding MSI. 
These factors have been ignored in image fusion algorithms proposed to date, and can have a significant impact on the solution of practical problems.

In this paper, we propose a new image fusion method accounting for spectral variability between the images, which can be caused by both acquisition (e.g. atmospheric, illumination) or seasonal changes.
Differently from existing approaches, we allow the high-resolution images underlying the observed HS and MS images to be different from one another. Employing a subspace/unmixing-based formulation, we use a unique set of endmembers for each image, employing a parametric model to represent the variability of the spectral signatures. The proposed algorithm, which is called \textit{HS-MS image Fusion with spectral Variability} (FuVar), estimates the subspace components (endmembers and abundance maps) of the high-resolution images using an alternating optimization approach, making use of the Alternating Direction Method of Multipliers to solve each subproblem. Simulation results with synthetic and real data show that the proposed approach performs similarly or better than state-of-the-art methods for images acquired under the same conditions, and much better when there is spectral variability between them.

The paper is organized as follows. In Section~\ref{sec:model_var}, the HS and MS observation process is presented, and a new model to represent spectral variability between these images is proposed. The proposed FuVar algorithm is introduced in Section~\ref{sec:prop_solution}. Experimental results with synthetic and real remote sensing images are presented in Section~\ref{sec:results}. Finally, Section~\ref{sec:conclusions} concludes the paper.





\section{Spectral Variability in Image Fusion} \label{sec:model_var}

%


Let $\bY_{\!h}\in\amsmathbb{R}^{L_h\times N}$ be an observed low spatial resolution HSI with $L_h$ bands and $N$ pixels, and $\bY_{\!m}\in\amsmathbb{R}^{L_m\times M}$ an observed low spectral resolution MSI with $L_m$ bands and $M$ pixels, with $L_m<L_h$ and $N<M$.
The image fusion problem consists of estimating an underlying image $\bZ\in\amsmathbb{R}^{L_h\times M}$ with high spatial and spectral resolutions, given $\bY_{\!h}$ and $\bY_{\!m}$.

A common approach to solve this problem is based on spectral unmixing using the linear mixture model (LMM).
The LMM assumes that the spectral representation of an observed image can be decomposed as a convex combination of a small number $P\ll L_h$ of spectral signatures (called endmembers) of pure materials in the scene~\cite{Keshava:2002p5667}. Let
\begin{equation} \label{eq:LMMforZ}
 \bZ = \bM\bA
\end{equation}
be the LMM for the underlying image $\bZ$. In~\eqref{eq:LMMforZ}, the columns of $\bM\in\amsmathbb{R}^{L_h\times P}$ are the spectral signatures of the $P$ endmembers, and each column of $\bA\in\amsmathbb{R}^{P\times M}$ is the vector of fractional abundances for the corresponding pixel of~$\bZ$.

%
%
%

Unmixing-based formulations successfully exploit the reduced dimensionality of the problem since only matrices $\bM$ and $\bA$ (which are much smaller than $\bZ$) need to be estimated.
Furthermore, a good estimate of $\bM$ for the fusion problem can be obtained directly from $\bY_{\!h}$ due to its high spectral resolution~\cite{simoes2015HySure}.
Other subspace estimation techniques, such as SVD or PCA, can also be employed to represent~$\bZ$. However, the LMM is usually preferred for its connection with the physical mixing process and for the better performance empirically verified in practice~\cite{yokoya2017HS_MS_fusinoRev,simoes2015HySure}.

The HS-MS image fusion literature assumes that all observed images are acquired at exactly the same conditions. This consideration allows both the HS and the MS images to be directly related to a unique underlying image~$\bZ$. This, however, is a major limitation of existing methodologies since a vast amount of data available for processing may come from sensors on board of different instruments or missions (e.g. AVIRIS and Sentinel images). In such scenarios, the acquisition conditions can be very different for HS and MS images, especially when there is a significant time interval between their capture.
Such situation is commonly encountered when data provided by several different MS sensors is used to generate high temporal resolution sequences.
%
The use of different sensors introduces variability in the spectral signatures of the observed images due to, for instance, varying illumination, different atmospheric conditions, or seasonal spectral variability of the materials in the scene (e.g. in the case of vegetation analysis)~\cite{Zare-2014-ID324,Somers:2009p6577}.
These effects cannot be adequately modeled by unmixing or subspace formulations that assume a single underlying image~$\bZ$ for both the HS and MS images. These methods assume that the underlying materials $\bM$ have the same spectral signatures in both images, and thus can be decomposed using the same subspace components~\cite{yokoya2012coupledNMF,simoes2015HySure,lanaras2015coupledUnmixingFusion}.  In the following, we propose a new model that accounts for the effects of spectral variability in HS-MS image fusion.

\subsection{The proposed model}

Consider distinct high-resolution images~$\bZ_h$ and~$\bZ_m$, both with $L_h$ bands and $M$ pixels, corresponding to the observed HSI and MSI, respectively.  These high resolution (HR) images can be different due to seasonal variability effects. We then decompose $\bZ_h$ and~$\bZ_m$ as
\begin{align} \label{eq:model_prop_i}
	\bZ_h {}={} \bM_h \bA \,, \quad
    \bZ_m {}={} \bM_m \bA
\end{align}
where $\bA\in\amsmathbb{R}^{P\times M}$ is the fractional abundance matrix, assumed common to both images, and~$\bM_h$, $\bM_m$ $\in\amsmathbb{R}^{L_h\times P}$ are the endmember matrices of the HSI and MSI, respectively. $\bM_h$ and $\bM_m$ can be different to account for spectral variability. For simplicity we have assumed the number of endmembers $P$ to be the same for both the hyperspectral and multispectral images. However, the materials represented in $\bM_h$ and $\bM_m$ can be completely different from one another.

Using~\eqref{eq:model_prop_i}, we assume that the observed HSI and MSI are generated according to
\begin{align}
\begin{split}\label{eq:observation_model}
	\bY_{\!h} {}={} & \bM_h \bA \bF \bD + \bN_h
    \\
    \bY_{\!m} {}={} & \bR \bM_m \bA + \bN_m
\end{split}
\end{align}
where $\bF\in\amsmathbb{R}^{M\times M}$ accounts for optical blurring due to the sensor point spread function, $\bD\in\amsmathbb{R}^{M\times N}$ is a spatial downsampling matrix, $\bR\in\amsmathbb{R}^{L_m\times L_h}$ is a matrix containing the spectral response functions (SRF) of each band of the MS instrument. $\bN_h\in\amsmathbb{R}^{L_h\times N}$ and $\bN_m\in\amsmathbb{R}^{L_m\times M}$ represent additive noise.

Matrix $\bM_h$ can be estimated from the observed HSI $\bY_{\!h}$ using endmember extraction or subspace decomposition. However, the same is not true for the estimation of $\bM_m$ since the MSI~$\bY_{\!m}$ has low spectral resolution. To address this issue, we propose to write $\bM_m$ as a function of $\bM_h$ using a specific model for spectral variability. 

Different parametric models have been recently proposed to account for spectral variability in hyperspectral unmixing. The extended  LMM proposed in~\cite{drumetz2016blindUnmixingELMMvariability} allows the scaling of each endmember spectral signature by a constant. This model effectively represents illumination and topographic variations, but is limited in capturing more complex variability effects. The perturbed LMM proposed in~\cite{thouvenin2016hyperspectralPLMM} adds an arbitrary variation matrix to a reference endmember matrix. This model lacks a clear connection to the underlying physical phenomena. The generalized LMM (GLMM), recently proposed in~\cite{imbiriba2018glmm}, accounts for more complex spectral variability effects by considering an individual scaling factor for each spectral band of the endmembers. This model introduces a connection between the amount of spectral variability and the amplitude of the reference spectral signatures, agreeing with practical observations.

Considering the GLMM, we propose to model the multispectral endmember matrix $\bM_m$ as a function of the endmembers extracted from the HSI as
\begin{align}
	\bM_m {}={} \bPsi \circ \bM_h
\end{align}
where $\bPsi\in\amsmathbb{R}^{L_h\times P}$ is a matrix of positive scaling factors and~$\circ$ denotes the Hadamard product. This model can represent spectral variability caused by seasonal changes~\cite{Zare-2014-ID324,Somers:2009p6577}, global illumination~\cite{drumetz2016blindUnmixingELMMvariability,drumetz2019hapkeELMM} or global atmospheric variations~\cite{shaw2003spectralImagRemote,uezato2016bundles}.
Then, the image fusion problem can finally be formulated as the problem of recovering the matrices $\bM_h$, $\bPsi$ and $\bA$ from the observed HS and MS images $\bY_{\!h}$ and $\bY_{\!m}$.




\section{The image fusion problem} \label{sec:prop_solution}


Considering the observation model in~\eqref{eq:observation_model} and assuming that matrices $\bM_h$, $\bF$, $\bD$ and $\bR$ are known or estimated previously, the image fusion problem reduces to estimating the variability and abundance matrices $\bPsi$ and $\bA$. Therefore, we propose to solve the image fusion problem through the following optimization problem:
%
%
\begin{align} \label{eq:opt_main_i}
	\bA^*,\bPsi^* {}={} & 
    \underset{\bA\geq0,\,\bPsi\geq0}{\arg\min}  \,\,\, 
    \frac{1}{2}\|\bY_{\!h} - \bM_h\bA\bF\bD\|_F^2
    \nonumber \\ & \quad
    + \frac{1}{2}\|\bY_{\!m} - \bR(\bPsi\circ\bM_h)\bA\|_F^2
    + \lambda_A \mathcal{R}(\bA)
    \nonumber \\ & \quad 
	+ \frac{\lambda_1}{2} 
    \|\bPsi - \cb{1}_L\cb{1}_P^\top \|_F^2
    + \frac{\lambda_2}{2}\|\bH_{\!\ell}\bPsi \|_F^2
\end{align}
where parameters $\lambda_1$, $\lambda_2$ and $\lambda_A$ balance the contributions of the different regularization terms to the cost function. The last two terms are regularizations over $\bPsi$. The first of them controls the amount of spectral variability by constraining $\bPsi$ to be close to unity. The second enforces spectral smoothness by making $\bH_{\!\ell}$ a differential operator. $\mathcal{R}(\bA)$ is a regularization over $\bA$ to enforce spatial smoothness, and is given by
\begin{align} \label{eq:regularization_term_A}
	\mathcal{R}(\bA) {}={} & \| \mathcal{H}_h(\bA)\|_{2,1} + \| \mathcal{H}_v(\bA)\|_{2,1}
\end{align}
where the linear operators $\mathcal{H}_h$ and $\mathcal{H}_v$ compute the first-order horizontal and vertical gradients of a bidimensional signal, acting separately for each material of $\bA$.
The spatial regularization in the abundances is promoted by a mixed $\mathcal{L}_{2,1}$ norm of their gradient, where $\|\bX\|_{2,1}=\sum_{n=1}^N\|\bx_n\|_2$. This norm is used to promote sparsity of the gradient across different materials (i.e. to force neighboring pixels to be homogeneous in all constituent endmembers). The $\mathcal{L}_1$ norm can also be used, leading to the Total Variation regularization, where $\|\bX\|_{1}=\sum_{n=1}^N\|\bx_n\|_1$~\cite{iordache2012total}.

Although the problem defined in~\eqref{eq:opt_main_i} is non-convex with respect to both $\bA$ and $\bPsi$, it is convex with respect to each of the variables individually. Thus, we propose to use an alternating least squares strategy by minimizing the cost function iteratively with respect to each variable in order to find a local stationary point of~\eqref{eq:opt_main_i}.
This solution is presented in Algorithm~\ref{alg:proposed_alg}, and the individual steps are detailed in the following subsections.

\begin{algorithm} [bth]
\small
\SetKwInOut{Input}{Input}
\SetKwInOut{Output}{Output}
\caption{FuVar~\label{alg:proposed_alg}}
\Input{$\bY_{\!m}$, $\bY_{\!h}$, $\bF$, $\bD$, $\bR$ parameters $P$, $\lambda_A$, $\lambda_1$, $\lambda_2$.}
\Output{The estimated high-resolution images $\bZhat_h$ and~$\bZhat_m$.}
Estimate $\bM_h$ from $\bY_{\!h}$ using an endmember extraction algorithm\;
Initialize $\bPsi^{(0)}=\cb{1}_L\cb{1}_P^\top$ \; 
Initialize $\bA$ with a bicubic interpolation of the FCLS estimative \;
Set $i=0$ \;
\While{stopping criterion is not satisfied}{
$i=i+1$ \;
Compute $\bA^{(i)}$ by solving~\eqref{eq:als_subprob_a_ii} using $\bPsi\equiv\bPsi^{(i-1)}$\;
Compute $\bPsi^{(i)}$ by solving~\eqref{eq:opt_bpsi_i} using $\bA\equiv\bA^{(i)}$ \;
}
\KwRet $\bZhat_h=\bM_h\bA^{(i)}$, $\bZhat_{m}=(\bPsi^{(i)}\circ\bM_h)\bA^{(i)}$ \;
\end{algorithm}

\subsection{Optimizing w.r.t. \texorpdfstring{$\boldsymbol{A}$}{A}}\label{sec:opt_A}

To solve problem~\eqref{eq:opt_main_i} with respect to $\bA$ for a fixed $\bPsi$, we define the following optimization problem rewriting the terms in~\eqref{eq:opt_main_i} that depend on $\bA$: 
%
%
\begin{align} \label{eq:als_subprob_a_ii}
	\bA^* {}={} & 
    \underset{\bA}{\arg\min}  \,\,\, 
    \frac{1}{2}\|\bY_{\!h} - \bM_h\bA\bF\bD\|_F^2
    + \iota_{\bbR_+}(\bA) 
	\nonumber \\ & \quad
    + \frac{1}{2}\|\bY_{\!m} - \bR(\bPsi\circ\bM_h)\bA\|_F^2
    \nonumber \\ & \quad
    + \lambda_A \big(\|\mathcal{H}_h(\bA)\|_{2,1} 
    + \|\mathcal{H}_v(\bA)\|_{2,1}\big)
\end{align}
where $\iota_{\amsmathbb{R}_+}(\cdot)$ is the indicator function of $\amsmathbb{R}_+$ (i.e. $\iota_{\amsmathbb{R}_+}(a)=0$ if $a\geq0$ and $\iota_{\amsmathbb{R}_+}(a)=\infty$ if $a<0$) acting component-wise on its input, and enforces the abundance positivity constraint.

Since problem~\eqref{eq:als_subprob_a_ii} is not separable w.r.t. the pixels neither differentiable due to the presence of the $\mathcal{L}_{2,1}$ norm we resort to a distributed optimization strategy, namely the alternating direction method of multipliers (ADMM). Specifically, the cost function in \eqref{eq:als_subprob_a_ii} must be represented in the following form~\cite{boyd2011ADMM}
\begin{align} \label{eq:admm_th_i}
	\min_{\bx,\bw} \left\{ f(\bx)+g(\bw) \, \Big|
    \mathcal{A}\bx+\mathcal{B}\bw=\bc
    \right\} ,
\end{align}
where $f:\amsmathbb{R}^a\to\amsmathbb{R}_+$ and $g:\amsmathbb{R}^b\to\amsmathbb{R}_+$ are closed, proper and convex functions, and $\bx\in\amsmathbb{R}^a$, $\bw\in\amsmathbb{R}^b$, $\bc\in\amsmathbb{R}^c$, $\mathcal{A}\in\amsmathbb{R}^{c\times a}$, $\mathcal{B}\in\amsmathbb{R}^{c\times b}$.

Using the scaled formulation of the ADMM~\cite[Sec. 3.1.1]{boyd2011ADMM}, the augmented Lagrangian associated with~\eqref{eq:admm_th_i} is given by
\begin{align} \label{eq:admm_th_ii}
	\mathcal{L}(\bx,\bw,\bu) {}={} f(\bx) + g(\bw) + \frac{\rho}{2} 
    \big\| \mathcal{A}\bx + \mathcal{B}\bw - \bc + \bu \big\|_2^2
\end{align}
with $\bu\in\amsmathbb{R}^{c}$ the scaled dual variable, and $\rho>0$.
The ADMM consists of updating the variables $\bx$, $\bw$ and $\bu$ sequentially. Denoting the primal and dual variables at the $k$-th iteration of the algorithm by $\bx^{(k)}$, $\bw^{(k)}$ and $\bu^{(k)}$, the ADMM can be expressed as
\begin{align} \label{eq:admm_th_iii}
\begin{split}
	\bx^{(k+1)} & {}\in{} \arg\min_{\bx} \mathcal{L}(\bx,\bw^{(k)},\bu^{(k)})
    \\
    \bw^{(k+1)} & {}\in{} \arg\min_{\bw} \mathcal{L}(\bx^{(k)},\bw,\bu^{(k)})
    \\
    \bu^{(k+1)} & {}={} \bu^{(k)} + \mathcal{A}\bx^{(k+1)} 
    + \mathcal{B}\bw^{(k+1)} - \bc .
\end{split}
\end{align}

To represent optimization problem~\eqref{eq:als_subprob_a_ii} in the form of~\eqref{eq:admm_th_i}, we define the following variables:
\begin{equation} \label{eq:variab_change_admm_a_i}
\begin{split}
	\bB_1^\top &= \bD^\top \bF^\top \bB_2^\top
	\\ 
    \bB_2 &= \bM_h \bA
	\\ 
    \bB_3 &= \bA
\end{split}
\hspace{8ex}
\begin{split}
    & \bB_4 = \mathcal{H}_h(\bB_3)
	\\ &
    \bB_5 = \mathcal{H}_v(\bB_3)
	\\ &
    \bB_6 = \bA.
\end{split}
\end{equation}

The optimization problem~\eqref{eq:als_subprob_a_ii} is then rewritten as
%
\begin{align}
	\bA^* {}={} & \underset{\bA}{\arg\min}  
    \,\,\, 
    \frac{1}{2}\|\bY_{\!h} - \bB_1\|_F^2
    + \lambda_A \big(\|\bB_4\|_{2,1} + \|\bB_5\|_{2,1}\big)
    \nonumber \\ & \hspace{0.7cm}
	+ \frac{1}{2}\|\bY_{\!m} - \bR(\bPsi\circ\bM_h) \bA\|_F^2
    + \iota_+(\bB_6) \label{eq:admm_prob_1}
	\\ &
    \hspace{0.3cm} \text{subject to the equations in}~\eqref{eq:variab_change_admm_a_i}
    \nonumber
\end{align}
%
%
This problem is equivalent to \eqref{eq:admm_th_i}, with $\bc=\cb{0}$, $\bx$ given by~\eqref{eq:variab_change_admm_a_ii}, where $\emph{vec}(\cdot)$ is the vectorization operator, and $\bw$, $\mathcal{A}$ and $\mathcal{B}$ given as follows (Eqs.~\eqref{eq:variab_change_admm_a_iii}--\eqref{eq:variab_change_admm_a_vi})

\begin{figure*}
\begin{align} \label{eq:variab_change_admm_a_ii}
\bx = 
\left[\begin{array}{cccccccc}
\emph{vec}(\bB_1^\top)^\top & 
\emph{vec}(\bB_2^\top)^\top & 
\emph{vec}(\bB_3)^\top & 
\emph{vec}(\bB_4)^\top & 
\emph{vec}(\bB_5)^\top & 
\emph{vec}(\bB_6)^\top 
\end{array}\right]^\top
\end{align}
\end{figure*}

\begin{align} \label{eq:variab_change_admm_a_iii}
\bw = \emph{vec}(\bA)
\end{align}

\begin{align} \label{eq:variab_change_admm_a_v}
& \mathcal{A} = 
\left[\begin{array}{cccccccc}
-\bI & \emph{rep}_L(\bD^\top\bF^\top)  &  0  &  0 & 0 & 0 \\
0 & \bP_{\top} & 0 & 0 & 0 & 0 \\
0 & 0 & \bI & 0 & 0 & 0 \\
0 & 0 & -\bH_h & \bI & 0 & 0 \\
0 & 0 & -\bH_v & 0 & \bI & 0 \\
0 & 0 & 0 & 0 & 0 & \bI
\end{array}\right]
\end{align}
where $\bP_{\top}$ is a permutation matrix that performs the transposition of the vectorized HSI, that is, $\bP_{\top}\emph{vec}(\bB_2^\top)=\emph{vec}(\bB_2)$, and
\begin{align} \label{eq:variab_change_admm_a_vi}
\mathcal{B} = 
\Big[\begin{array}{cccccccc}
0 &
-\emph{rep}_M(\bM_h^\top) &
-\bI &
0 &
0 &
-\bI
\end{array}\Big]^ \top
\end{align}
where $\emph{rep}_M(\bB)$ are block diagonal matrices with~$M$ repeats of matrix $\bB$ in the main diagonal, that is, 
\begin{align}
\emph{rep}_M(\bB)  = 
\left[\begin{array}{cccccccc}
\bB & 0 &\cdots & 0 \\
0 & \bB & & 0 \\
\vdots & & \ddots & \vdots \\
0 & 0 & \cdots & \bB
\end{array}\right].
\end{align}

Using~\eqref{eq:variab_change_admm_a_iii}--\eqref{eq:variab_change_admm_a_vi} in~\eqref{eq:admm_th_ii}, the scaled augmented Lagrangian of~\eqref{eq:admm_prob_1} can be written as
\begin{align}  \label{eq:admm_lagrangean}
    \mathcal{L}({\bTheta}) {}={} &
    \frac{1}{2}\|\bY_{\!h} - \bB_1\|_F^2
    \nonumber \\ & 
	+ \frac{1}{2}\|\emph{vec}(\bY_{\!m}) 
    - \emph{rep}_M(\bR(\bPsi\circ\bM_h)) \bw\|^2
    \nonumber \\ & 
	+ \lambda_A \big(\|\bB_4\|_{2,1} + \|\bB_5\|_{2,1}\big)
    + \iota_+(\bB_6) 
	\nonumber \\ &
    + \frac{\rho}{2}\Big(
    \|\emph{rep}_L(\bD^\top\bF^\top)\emph{vec}(\bB_2^\top) 
    - \emph{vec}(\bB_1^\top) + \bu_1 \|^2
    \nonumber \\ & 
    + \|\emph{vec}(\bB_2) - \emph{rep}_M(\bM_h) \bw + \bu_2\|^2 
    \nonumber \\ & 
    + \|\emph{vec}(\bB_3) - \bw + \bu_3\|^2
    \nonumber \\ & 
    + \|\emph{vec}(\bB_4) - \bH_h \emph{vec}(\bB_3) + \bu_4\|^2
    \nonumber \\ & 
    + \|\emph{vec}(\bB_5) - \bH_v \emph{vec}(\bB_3) + \bu_5\|^2
    \nonumber \\ & 
    + \|\emph{vec}(\bB_6) - \bw + \bu_6\|^2 \Big)
\end{align}
where ${\bTheta}=\{\bw,\bB_1,\bB_2,\bB_3,\bB_4,\bB_5,\bB_6,\bu\}$ and $\bu_i$, $i=1,\ldots,6$ form the partition of the dual variable $\bu$ with compatible dimensions, which is given by
\begin{align} \label{eq:variab_change_admm_a_iv}
\bu {}={} 
\left[\begin{array}{cccccccc}
\bu_1^\top & \bu_2^\top & \bu_3^\top & \bu_4^\top 
& \bu_5^\top & \bu_6^\top 
\end{array}\right]^\top  \,.
\end{align}
The optimization of $\mathcal{L}(\bTheta)$ with respect to each of the variables in~$\bTheta$ is described in detail in Appendix~\ref{sec:appendix1}.

\subsection{Optimizing w.r.t. $\bPsi$}

The optimization problem with respect to $\bPsi$ and considering $\bA$ fixed can be recast from~\eqref{eq:opt_main_i} by considering only terms that depend on $\bPsi$:
%
\begin{align} \label{eq:opt_bpsi_i}
	\bPsi^* {}={} & 
    \underset{\bPsi\geq0}{\arg\min}  \,\,\, 
    \frac{1}{2}\|\bY_{\!m} - \bR(\bPsi\circ\bM_h)\bA\|_F^2
    \nonumber \\ &
	\quad + \frac{\lambda_1}{2} 
    \|\bPsi - \cb{1}_L\cb{1}_P^\top \|_F^2
    + \frac{\lambda_2}{2} \|\bH_{\!\ell}\bPsi\|_F^2
\end{align}

Analogously to Section~\ref{sec:opt_A}, the ADMM can be used to split problem~\eqref{eq:opt_bpsi_i} into smaller simpler problems that are solved iteratively. Specifically, we denote
\begin{align}
\begin{split}\label{eq:admm_vars_2}
	&
	\bB = \bPsi \circ \bM_h
    \\&
    \bx = \emph{vec}(\bB)
    \\&
    \bw {}={} \emph{vec}(\bPsi)
    \\&
    \mathcal{A} {}={} \bI
    \\&
    \mathcal{B} {}={} -\emph{diag}(\emph{vec}(\bM_h))\,.
\end{split}
\end{align}

Then, the optimization problem~\eqref{eq:opt_bpsi_i} can be expressed as
%
\begin{align} \label{eq:opt_bpsi_ii}
	\bPsi^* {}={} & 
    \underset{\bPsi}{\arg\min}  \,\,\, 
    \frac{1}{2}\|\bY_{\!m} - \bR\bB\bA\|_F^2
    + \iota_+(\bPsi)
    \nonumber \\ &
	\quad + \frac{\lambda_1}{2} 
    \|\bPsi - \cb{1}_L\cb{1}_P^\top \|_F^2
    + \frac{\lambda_2}{2} \|\bH_{\!\ell}\bPsi\|_F^2
    \\ & \nonumber
    \text{subject to } \bB = \bPsi \circ \bM_h
\end{align}
which is equivalent to the ADMM problem.

The scaled augmented Lagrangian for this problem can be written as
\begin{align} \label{eq:opt_bpsi_iii}
	\mathcal{L}(\bB,\bPsi,\bU) & {}={}  
    \frac{1}{2}\|\bY_{\!m} - \bR\bB\bA\|_F^2
    + \iota_+(\bPsi)
    \nonumber \\ & \quad 
    + \frac{\lambda_1}{2} \|\bPsi - \cb{1}_L\cb{1}_P^\top \|_F^2
    + \frac{\lambda_2}{2} \|\bH_{\!\ell}\bPsi\|_F^2
    \nonumber \\ & \quad
    + \frac{\rho}{2} \| \bB - \bPsi\circ\bM_h + \bU\|_F^2\,.
\end{align}
The optimization of $\mathcal{L}(\bB,\bPsi,\bU)$ with respect to each of the variables~$\bB$, $\bPsi$ and $\bU$ is described in detail in Appendix~\ref{sec:appendix2}.

\section{Experimental Results} \label{sec:results}

Three experiments were devised to illustrate the performance of the proposed method.
The first one uses synthetically generated images with controlled spectral variability in order to assess the estimation performance.
The second compares the performances of the proposed method with those of other techniques when there is no spectral variability between the images (i.e. acquired at the same time instant).
Finally, the third example depicts the fusion of real HS and MS images acquired at different time instants, thus containing spectral variability.
Details of the three examples and their experimental setups are described next.

\subsection{Experimental Setup}
\label{sec:exp_setup}


We compare the proposed method to three other techniques, namely: the unmixing-based HySure~\cite{simoes2015HySure} and the CNMF~\cite{yokoya2012coupledNMF} algorithms, and the multiresolution analysis-based GLP-HS algorithm~\cite{aiazzi2006GLP_HS}. These methods   provided the best overall fusion performance in extensive experiments reported in survey~\cite{yokoya2017HS_MS_fusinoRev}.

For experiments~2 and~3, which are based on real hyperspectral and multispectral images acquired at the same spatial resolution, two preprocessing steps were performed based on the same procedure described in~\cite{simoes2015HySure}. First, spectral bands with low SNR or corresponding to water absorption spectra were manually removed.
%
Next, all bands of the hyperspectral and multispectral images were normalized such that the 0.999 intensity quantile corresponded to a value of 1. 
Finally, the HSI was denoised using the method described in~\cite{roger1996denoisingHSI} to yield high-SNR (hyperspectral) reference images~$\bZ_h$~\cite{yokoya2017HS_MS_fusinoRev}.

For all experiments, the observed HSIs were generated by blurring the reference HSI image with a Gaussian filter with unity variance, decimating it and adding noise to obtain a 30dB SNR. The observed MSIs were obtained from the reference MSI image by adding noise to obtain a 40dB SNR.

We fixed the number of endmembers at $P=30$ for the CNMF, HySure and FuVar methods according to~\cite{yokoya2017HS_MS_fusinoRev}. The blurring kernel $\bF$ was assumed to be known a priori for the HySure and FuVar methods, and $\bM_h$ was estimated from the HSI using the VCA algorithm~\cite{nascimento2005vca}. The regularization parameters for the proposed method were empirically set at $\lambda_A=10^{-4}$, $\lambda_1=0.01$ and $\lambda_2=10^4$. We selected a large value for $\lambda_2$ to have spectrally smooth variability, and a small value of $\lambda_1$ to allow the scaling factors $\bPsi$ to adequately fit the data. Nevertheless, the proposed method showed to be fairly insensitive to the choice of parameters.
We ran the alternating optimization process in Algorithm~\ref{alg:proposed_alg} for at most~10 iterations or until the relative change of $\bA$ and $\bPsi$ was less than~$10^{-3}$.

The visual assessment of the reconstructed images is performed using one true-color image at the visible spectrum (with red, green and blue corresponding to $0.45$, $0.56$ and $0.66\mu m$) and one pseudo-color image at the infrared spectrum (with red, green and blue corresponding to $0.80$, $1.50$ and~$2.20\mu m$).

\subsection{Quality measures:}

We use four quality metrics previously considered in~\mbox{\cite{yokoya2017HS_MS_fusinoRev,simoes2015HySure}} to evaluate the quality of the reconstructed images, denoted by~$\bZhat$, by comparing them to the original HR images~$\bZ$.
The first one is the peak signal to noise ratio (PSNR), computed bandwise and defined as
\begin{align}
	PSNR(\bZ,\bZhat) {}={} \frac{1}{L} \sum_{\ell=1}^L
    10 \log_{10} \bigg(
    \frac{M\,\Ex\{\max(\bZ_{\ell,:})\}}{\|\bZ_{\ell,:}
    -\bZhat_{\ell,:}\|_F^2} \bigg)
\end{align}
where $\Ex\{\cdot\}$ is the expected value operator and $\bZ_{\ell,:}\in\amsmathbb{R}^{1\times M}$ denotes the $\ell$-th band of $\bZ$. Larger PSNR values indicate higher quality of spatial reconstruction, tending to infinity when the bandwise reconstruction error tends to zero.

The second metric is the Spectral Angle Mapper (SAM):
\begin{align}
	SAM(\bZ,\bZhat) {}={} \frac{1}{M} \sum_{n=1}^M 
    \cos^{-1}\bigg(\frac{\bZ_{:,n}^\top\bZhat_{:,n}}
    {\|\bZ_{:,n}\|_2\|\bZhat_{:,n}\|_2} \bigg)
    \,.
\end{align}

The third metric is the \textit{Erreur Relative Globale Adimensionnelle de Synthèse} (ERGAS)~\cite{wald2000qualityERGAS}, which provides a global statistical measure of the quality of the fused data with the best value at 0:
\begin{align}
	ERGAS(\bZ,\bZhat) {}={} 100\sqrt{\frac{N}{L\,M}
    \sum_{\ell=1}^L \frac{\|\bZ_{\ell,:}
    -\bZhat_{\ell,:}\|_F^2}{\emph{mean}(\bZhat_{\ell,:})^2}}
\end{align}
where $\emph{mean}(\cdot)$ computes the sample mean value of a signal.

Finally, the fourth metric is the Universal Image Quality Index (UIQI)~\cite{wang2002qualityUIQI}, extended to multiband images as
\begin{align}
	UIQI(\bZ,\bZhat) & {}={} \frac{4}{LK}\sum_{\ell=1}^L\sum_{k=1}^K
    \frac{\emph{cov}(\bP_k\bZ_{\ell,:},\bP_k\bZhat_{\ell,:})
    }{\emph{var}(\bP_k\bZ_{\ell,:}) + \emph{var}(\bP_k\bZhat_{\ell,:})}
    \nonumber \\ & \times
    \frac{\emph{mean}(\bP_k\bZ_{\ell,:})\emph{mean}(\bP_k\bZhat_{\ell,:})
    }{\emph{mean}(\bP_k\bZ_{\ell,:})^2 
    + \emph{mean}(\bP_k\bZhat_{\ell,:})^2}
\end{align}
where $\emph{var}(\cdot)$ computes the sample variance of a signal, $\emph{cov}(\bB_1,\bB_2)$ computes the sample covariance between $\bB_1$ and $\bB_2$, and $\bP_k$ is a projection matrix which extracts the $k$-th patch with $32\times32$ pixels from the images $\bZ_{\ell,:}$ and $\bZhat_{\ell,:}$, with $K$ being the total number of image patches. The UIQI is restricted to the interval~$[-1,1]$, corresponding to a perfect reconstruction when equal to one.

\subsection{Example 1}

This example was designed to evaluate the algorithms in a controlled environment. The true abundance matrix $\bA$, with $100\times100$ pixels and containing spatial correlation was generated using a Gaussian Random Fields algorithm. Three endmembers with $L=244$ bands were selected from the USGS spectral library and used to construct matrix~$\bM_h$.
The MS endmember matrix $\bM_m$ was obtained by multiplying $\bM_h$ with a matrix of scaling factors $\bPsi$ to introduce spectral variability. Matrix $\bPsi$ was constructed by random piecewise linear functions centered around a unitary gain.
The high-resolution HS and MS images $\bZ_h$ and $\bZ_m$ were generated from this data using the linear mixing model. 
The MS image $\bY_{\!m}$ with $L_m=16$ bands was then generated by applying a piecewise uniform SRF to $\bZ_m$. The low-resolution HS $\bY_{\!h}$ image was generated by blurring $\bZ_h$ with a Gaussian filter with unitary variance and decimating it by a factor of~$4$.
Finally, white Gaussian noise was added to both images, resulting in SNRs of 30dB for $\bY_{h}$ and 40dB for $\bY_{\!m}$.
Both the blurring kernel $\bF$ and the SRF $\bR$ were assumed to be known a priori for all algorithms. 

\begin{figure}[thb]
	\centering
	\includegraphics[width=\linewidth,trim={0cm 0cm 0cm 0cm},clip]{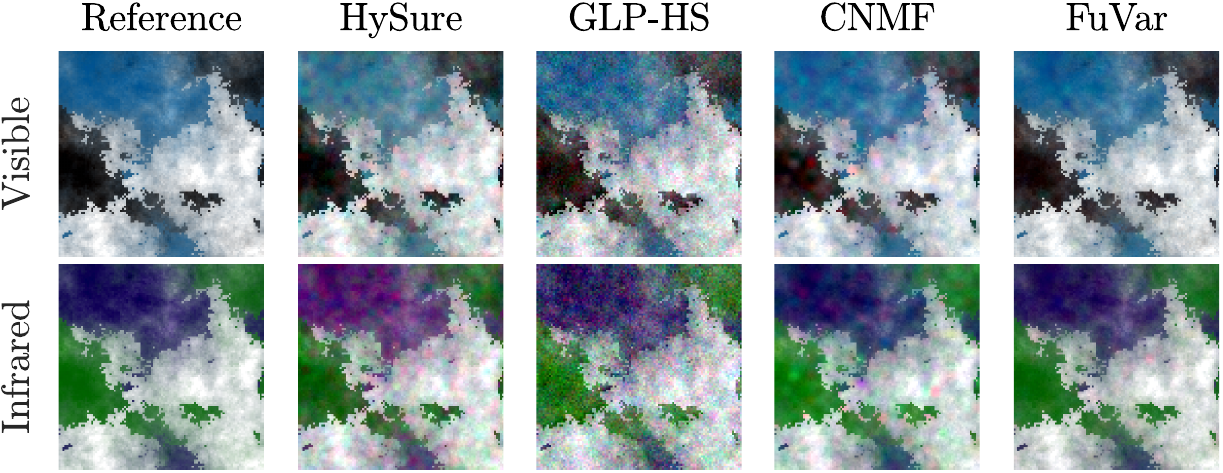}
    \caption{Visual results for the hyperspectral images in Example~1.}
    \label{fig:ex1_hsi_visual}
\end{figure}

\begin{figure}[h]
	\centering
	\includegraphics[width=\linewidth]{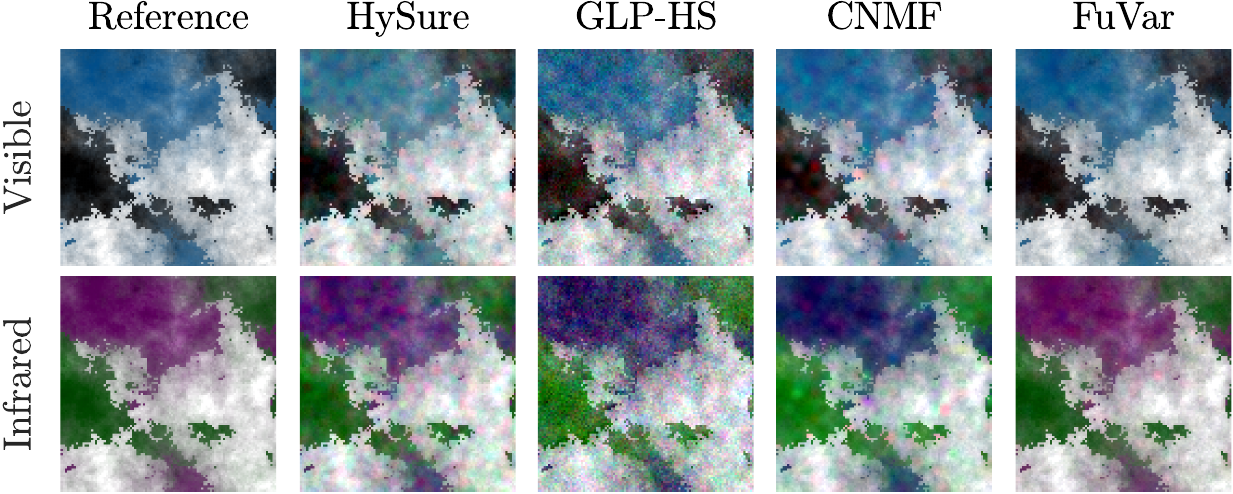}
    \caption{Visual results for the multispectral images in Example~1.}
    \label{fig:ex1_msi_visual}
\end{figure}

The quantitative results for the quality of the reconstructed images with respect to both $\bZ_h$ and $\bZ_m$ are shown in Table~\ref{tab:ex1_res}.
Figs.~\ref{fig:ex1_hsi_visual} and~\ref{fig:ex1_msi_visual} provide a visual assessment of the results for the super resolved HS and MS images $\bZ_h$ and $\bZ_m$.

\begin{table} [htb]
\caption{Example~1 - Results for the synthetic HS and MS images.}
\vspace{-0.15cm}
\centering
\renewcommand{\arraystretch}{1.3}
\setlength\tabcolsep{3.5pt}
\resizebox{\linewidth}{!}{%
\begin{tabular}{l|cccc|ccccccc}
\hline
& \multicolumn{4}{c|}{Hyperspectral HR image} & \multicolumn{4}{c}{Multispectral HR image} \\
\hline
& PSNR & SAM & ERGAS & UIQI & PSNR & SAM & ERGAS & UIQI \\
\hline
GLP-HS & 29.21 & 2.055 & 1.152 & 0.826 & 25.51 & 2.936 & 1.665 & 0.778\\
HySure & 31.00 & 1.430 & 1.013 & 0.936 & 33.56 & 1.611 & 0.761 & 0.942\\
CNMF   & 36.60 & 0.825 & 0.480 & 0.956 & 28.49 & 2.191 & 1.296 & 0.906\\
FuVar  & \textbf{42.22} & \textbf{0.509} & \textbf{0.257} & \textbf{0.989} & \textbf{42.69} & \textbf{0.497} & \textbf{0.242} & \textbf{0.980}\\
\hline
\end{tabular}}
\label{tab:ex1_res}
\end{table}

\subsubsection{Discussion}
%
The results presented in Table~\ref{tab:ex1_res} clearly show the performance gains obtained in all metrics using the proposed FuVar method. In terms of PSNR, the FuVar algorithm yielded gains larger than 5dB and 9dB for $\bZ_h$ and $\bZ_m$, respectively. 
Substantial performance improvements were also obtained for all other metrics, indicating that FuVar yields more accurate spatial and spectral results, resulting in better image reconstruction.
%
Visual inspection of Figs.~\ref{fig:ex1_hsi_visual} and~\ref{fig:ex1_msi_visual} shows that all methods yield reasonable results for both the visible and infrared regions of the spectrum. However, the most accurate reconstruction is clearly provided by FuVar, which keeps a good trade-off between the smooth parts and sharp discontinuities in the image without introducing spatial artifacts or color distortions. Such defects can be seen in the results from all other methods such as the CNMF and, especially, the HySure algorithm.

\subsection{Example 2}

This example compares the performance of the different methods when there is no spectral variability between the HS and MS images. We consider a dataset consisting of HS and MS images acquired above Paris with a spatial resolution of~30m. These images were initially presented in~\cite{simoes2015HySure} and captured by 
the Hyperion and the Advanced Land Imager instruments on board the Earth Observing-1 Mission satellite.

The reference HSI, with $72\times72$ pixels and $L=128$ bands was blurred by a Gaussian filter with unity variance and decimated by a factor of~$3$. Finally, WGN was added to generate the observed HSI~$\bY_{\!h}$ with a 30dB SNR. The observed MSI $\bY_{\!m}$ with $L_m=9$ bands was obtained from the reference MSI by adding WGN to achieve a 40dB SNR. The spectral response function $\bR$ was estimated from the observed images using the method described in~\cite{simoes2015HySure}.

\begin{figure}[thb]
	\centering
	\includegraphics[width=\linewidth,trim={0cm 0cm 0cm 0cm},clip]{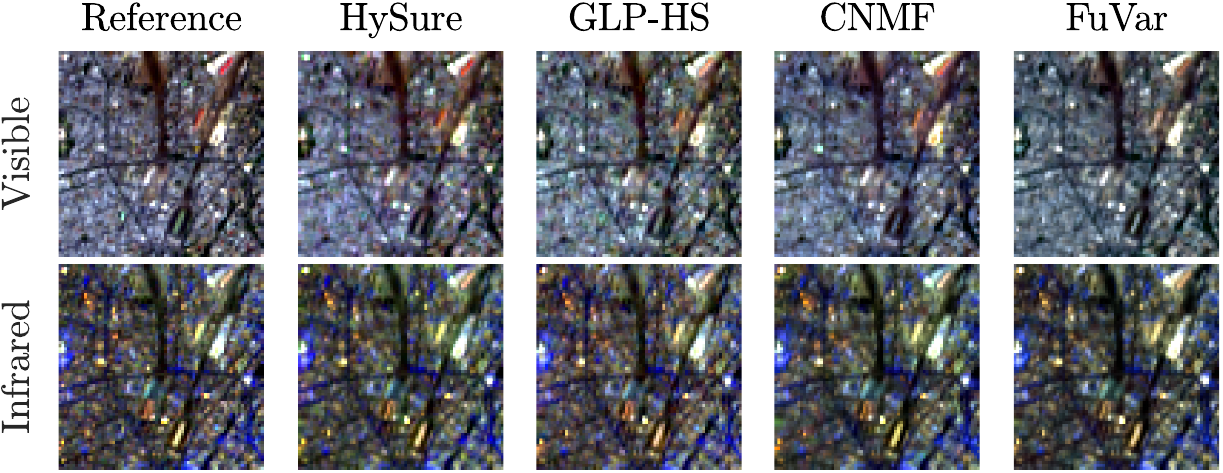}
    \caption{Visual results for the hyperspectral images in Example~2.}
    \label{fig:ex2_hsi_visual}
\end{figure}

%
The quantitative results for the quality of the reconstructed images with respect to~$\bZ_h$ are displayed in Table~\ref{tab:ex2_res}.
A visual assessment of the results for the super-resolved HS image $\bZ_h$ is presented in Fig.~\ref{fig:ex2_hsi_visual}.

\begin{table} [htb]
\caption{Example~2 - Results for the Paris image.}
\vspace{-0.15cm}
\centering
\renewcommand{\arraystretch}{1.3}
\setlength\tabcolsep{3.5pt}
{\begin{tabular}{l|cccc}
\hline
& PSNR & SAM & ERGAS & UIQI \\
\hline
GLP-HS & 26.78 & 3.811 & 5.218 & 0.778 \\
HySure & 28.78 & \textbf{3.060} & 4.169 & 0.844 \\
CNMF   & 28.08 & 3.432 & 4.529 & 0.818 \\
FuVar  & \textbf{28.97} & 3.156 & \textbf{4.016} & \textbf{0.848} \\
\hline
\end{tabular}}
\label{tab:ex2_res}
\end{table}

\subsubsection{Discussion}

Although the proposed method estimates the matrix $\bPsi$ to cope with the spectral variability often existing between the HS and MS images, its performance was still comparable with state of the art image fusion methods when there was no spectral variability. In fact, the results in Table~\ref{tab:ex2_res} show that FuVar yielded slightly better than the competing methods for all metrics but the SAM, for which it yielded the second best and very close to the best result. 
%
The slight improvement obtained by the FuVar algorithm can be attributed to the fact that the matrix $\bPsi$, originally devised to capture spectral variability, can also compensate for errors introduced by the estimation of the spectral response function~$\bR$.
Visually, the images displayed in Fig.~\ref{fig:ex2_hsi_visual} also indicate that the results obtained with all methods are very similar. This reconstruction similarity is in agreement with the quantitative results presented in Table~\ref{tab:ex2_res}.

\begin{figure}[thb]
	\centering
\begin{minipage}{0.49\linewidth}
        \centering
        \includegraphics[width=1\linewidth]{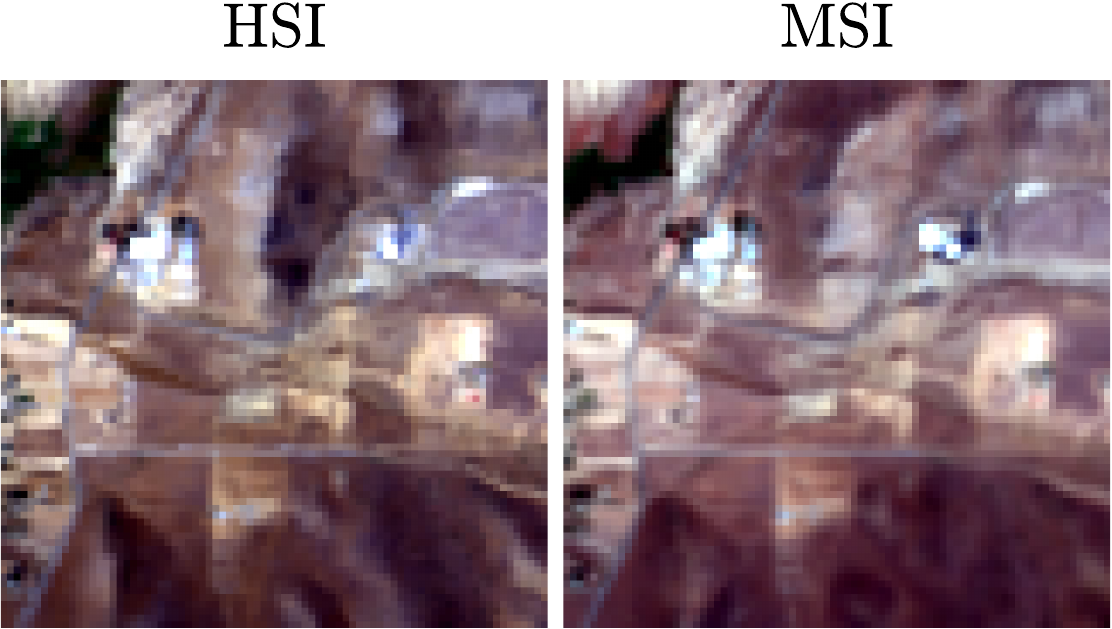}
        Lake Isabella
    \end{minipage}
    \begin{minipage}{0.49\linewidth}
        \centering
        \includegraphics[width=1\linewidth]{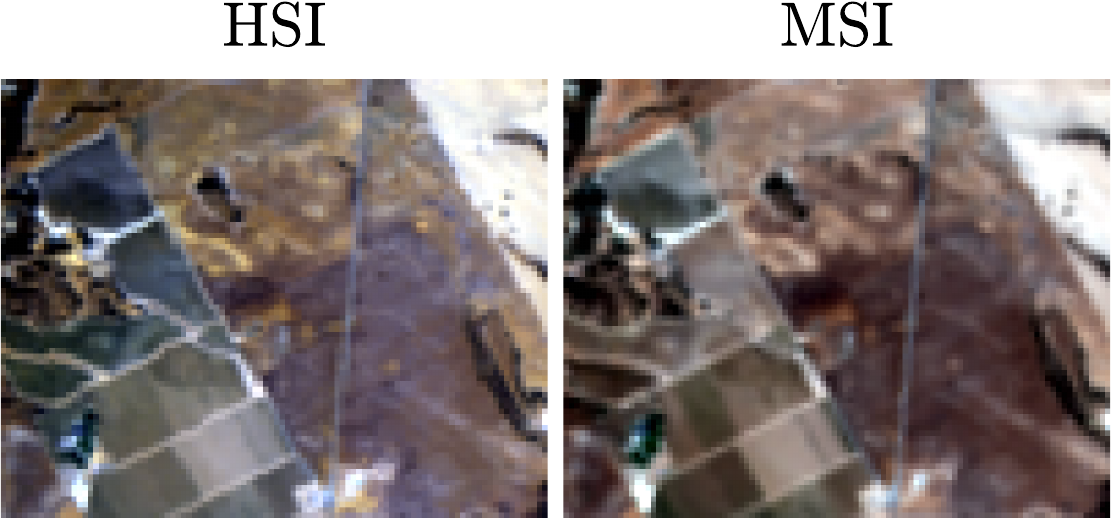}
        Lockwood
    \end{minipage}
    \caption{Hyperspectral and multispectral images with a small acquisition time difference used in Experiment~3.}
    \label{fig:ex3_HSIs_MSIs_a}
\end{figure}

\begin{figure}[thb]
	\centering
\begin{minipage}{0.48\linewidth}
        \centering
        \includegraphics[width=1\linewidth]{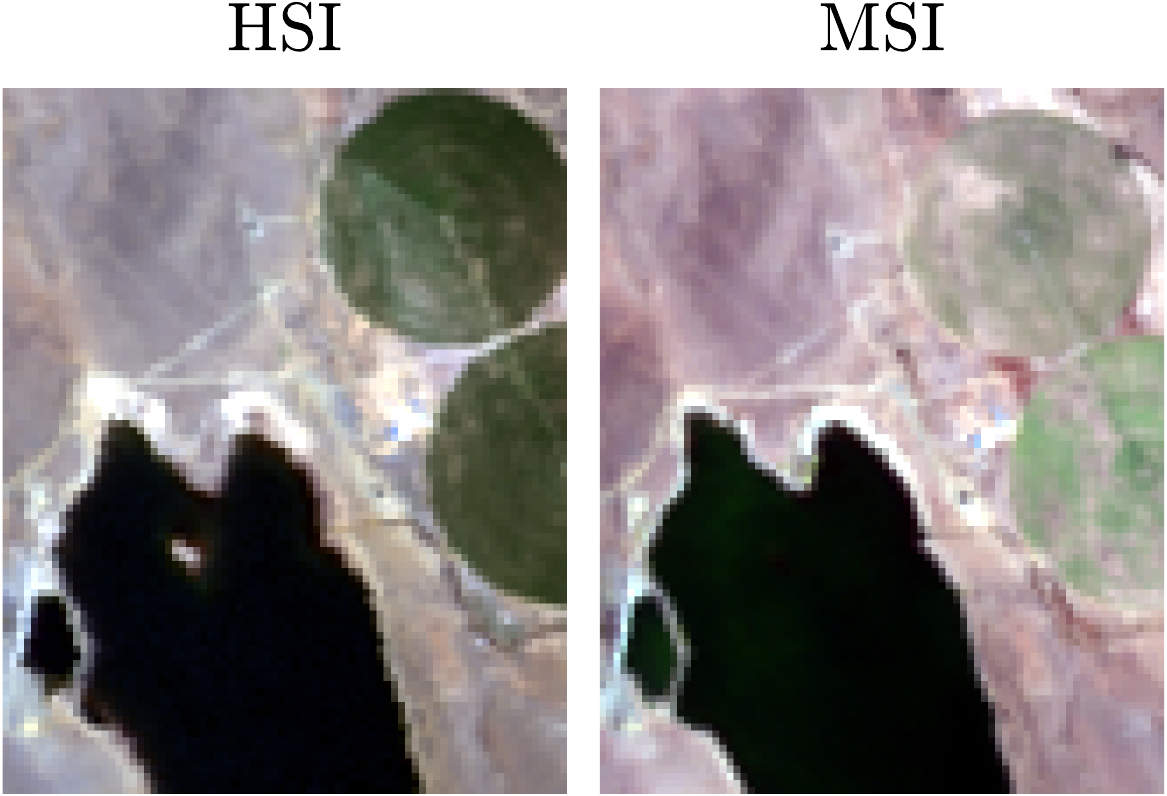}
        Lake Tahoe
    \end{minipage}
    \,\,
    \begin{minipage}{0.48\linewidth}
        \centering
        \includegraphics[width=1\linewidth]{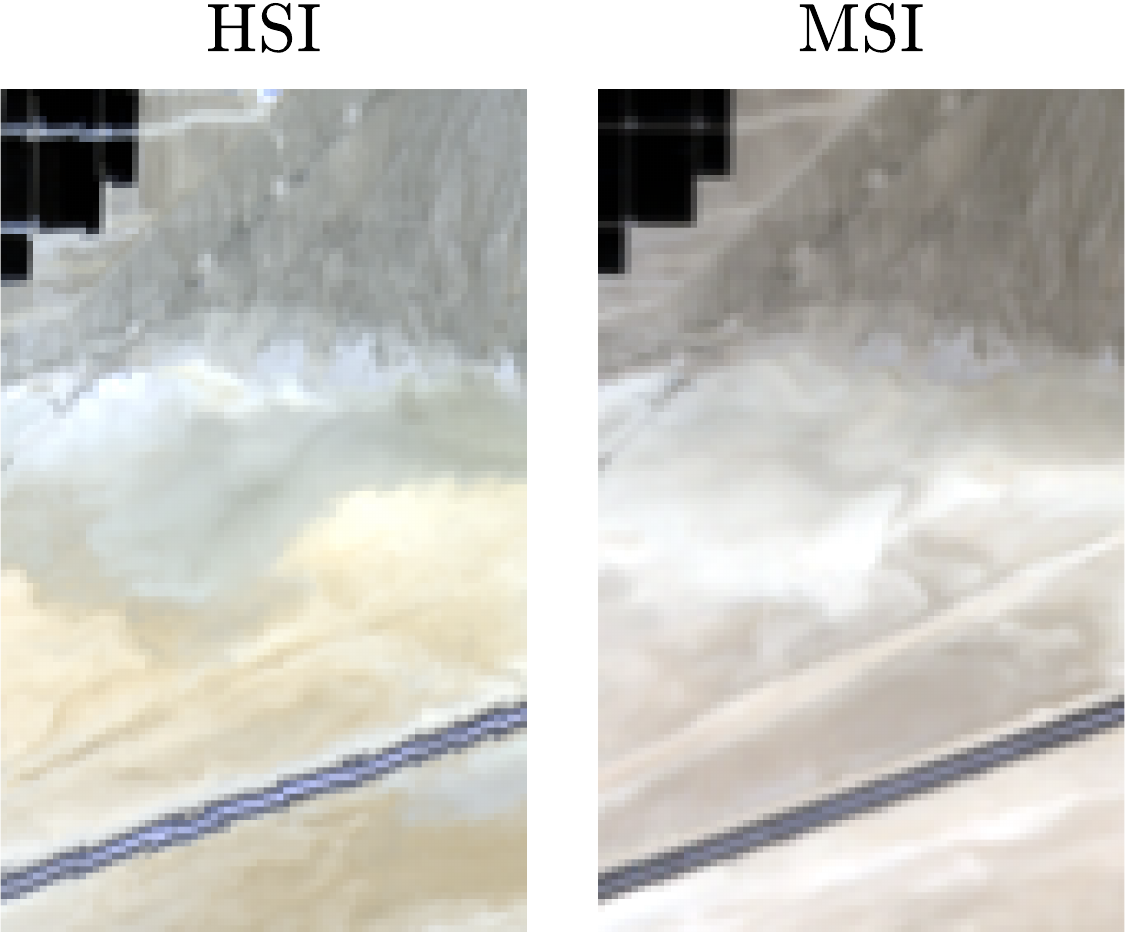}
        Ivanpah Playa
    \end{minipage}
    \caption{Hyperspectral and multispectral images with a large acquisition time difference used in Experiment~3.}
    \label{fig:ex3_HSIs_MSIs_b}
\end{figure}

\subsection{Example 3}

This example illustrates the performance of the proposed method when fusing HS and MS images acquired at different time instants, thus presenting seasonal spectral variability.
We consider four pairs of reference hyperspectral and multispectral images with a spatial resolution of~20m. The HS images were acquired by the AVIRIS instrument and contained $L=173$ bands after pre-processing, while the MS images containing $L_m=10$ bands was acquired by the Sentinel-2A instrument.

We divided this experiment in two parts using four different pairs of images\footnote{All HS and MS images used in this experiment are available at \url{https://aviris.jpl.nasa.gov/alt_locator/} and \url{https://earthexplorer.usgs.gov}.}. The first two pairs of images were acquired with a relatively small time difference between the HS and MS images (smaller than three months), and illustrate the performance of the methods for small spectral variability. The other two pairs of images were acquired with a larger time difference (more than one year), and illustrate a scenario with considerable spectral variability and seasonal changes.

In all cases, the observed image pairs $\bY_{\!h}$ and $\bY_{\!m}$ were generated as follows. The reference hyperspectral images were blurred by a Gaussian filter with unity variance and decimated by a factor of~$4$. Finally, WGN was added to generate the observed HSIs~$\bY_{\!h}$ with a 30dB SNR. The observed MS images $\bY_{\!m}$ was obtained by adding WGN to the reference MSI to obtain a 40dB SNR. The spectral response function $\bR$ obtained from calibration measurements and was thus known a priori\footnote{Available at https://earth.esa.int/web/sentinel/user-guides/sentinel-2-msi/document-library/-/asset\_publisher/Wk0TKajiISaR/content/sentinel-2a-spectral-responses}.

\subsubsection{HS and MS images with a small acquisition time difference}

The first pair of images, with $80\times80$ pixels, was acquired over the Lake Isabella area and can be seen in Fig.~\ref{fig:ex3_HSIs_MSIs_a}. The HS and MS images were acquired on 2018-06-27 and on 2018-08-27, respectively.
The second pair of images, with $80\times100$ pixels, was acquired near Lockwood and is also shown in Fig.~\ref{fig:ex3_HSIs_MSIs_a}. The HS and MS images were acquired on 2018-08-20 and on 2018-10-19, respectively. 
Although the HS and MS images are similar, there are slight differences in the color hues of the images. This is most clearly seen in the right part of the Lake Isabella image, and in the central ground stripe of the Lockwood image.

\begin{figure}[thb]
	\centering
	\includegraphics[width=\linewidth,trim={0cm 0cm 0cm 0cm},clip]{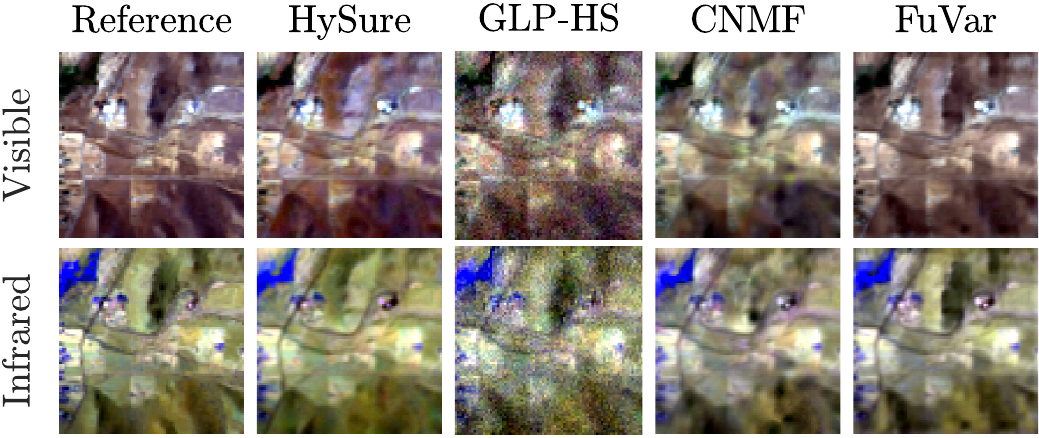}
    \caption{Visual results for the Lake Isabella hyperspectral image in Example~3.}
    \label{fig:ex3a_hsi_visual_a}
\end{figure}

\begin{figure}[thb]
	\centering
	\includegraphics[width=\linewidth,trim={0cm 0cm 0cm 0cm},clip]{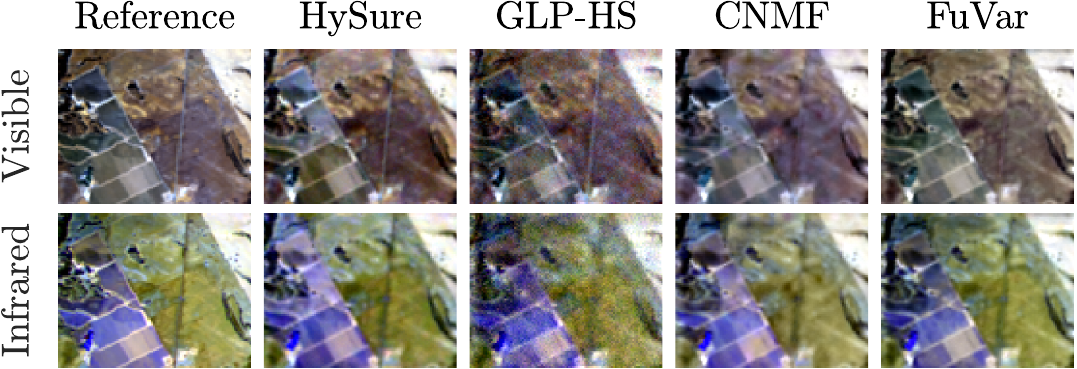}
    \caption{Visual results for the Lockwood hyperspectral image in Example~3.}
    \label{fig:ex3b_hsi_visual_a}
\end{figure}

The quantitative results for both images are displayed in Table~\ref{tab:ex3_res_a}, and a visual assessment of the results for the super-resolved HS images $\bZ_h$ is presented in Figs.~\ref{fig:ex3a_hsi_visual_a} and~\ref{fig:ex3b_hsi_visual_a}.

\begin{table} [htb]
\caption{Example~3 - Results for Lake Isabella and Lockwood images.}
\vspace{-0.15cm}
\centering
\renewcommand{\arraystretch}{1.3}
\setlength\tabcolsep{3.5pt}
\resizebox{\linewidth}{!}
{\begin{tabular}{l|cccc|cccccccccccccccc}
\hline
& \multicolumn{4}{c|}{Lake Isabella image} & \multicolumn{4}{c}{Lockwood image} \\
\hline
& PSNR & SAM & ERGAS & UIQI & PSNR & SAM & ERGAS & UIQI \\
\hline
GLP-HS & 25.23 & 2.981 & 3.510 & 0.813 & 25.10 & 2.819 & 3.146 & 0.776 \\
HySure & 20.08 & 2.932 & 5.160 & 0.653 & 23.17 & 2.416 & 3.497 & 0.770 \\
CNMF   & 25.30 & 2.261 & 3.517 & 0.793 & 25.62 & 2.207 & 2.870 & 0.786 \\
FuVar  & \textbf{28.80} & \textbf{2.249} & \textbf{3.036} & \textbf{0.932} & \textbf{27.87} & \textbf{2.131} & \textbf{2.838} & \textbf{0.893} \\
\hline
\end{tabular}}
\label{tab:ex3_res_a}
\end{table}


\paragraph{Discussion}

The results in Table~\ref{tab:ex3_res_a} show a superior performance of the FuVar algorithm when compared to those of the other algorithms. Although the SAM results for both images and the ERGAS results for the Lockwood image were similar between the FuVar and CNMF algorithms, the FuVar shows a considerable performance improvement in the remaining metrics for both data sets.
Visual inspection of the recovered super-resolved HSIs in Figs.~\ref{fig:ex3a_hsi_visual_a} and~\ref{fig:ex3b_hsi_visual_a} shows an accuracy improvement obtained by using FuVar over the competing algorithms for both the visible and infrared portions of the spectrum. Although the HySure algorithm resulted in coherent images without large color differences from the reference images in Fig.~\ref{fig:ex3a_hsi_visual_a}, slight differences can be found throughout the whole scene. The GLP-HS and CNMF showed even clearer color aberrations in the visible spectrum in Fig.~\ref{fig:ex3b_hsi_visual_a}. The FuVar algorithm results, on the other hand, are very close to the reference images for both examples.

\subsubsection{HS and MS images with a large acquisition time difference}

The third pair of images, with $100\times80$ pixels, was acquired over the Lake Tahoe area and can be seen in Fig.~\ref{fig:ex3_HSIs_MSIs_b}. The HS and MS images were acquired at 2014-10-04 and at 2017-10-24, respectively.
The fourth pair of images, with $80\times128$ pixels, was acquired over the Ivanpah Playa and are also shown in Fig.~\ref{fig:ex3_HSIs_MSIs_b}. The HS and MS images were acquired at 2015-10-26 and at 2017-12-17, respectively. 
A significant spectral variability between the images can be readily verified in both cases. For the Lake Tahoe image, it is evidenced by the different color hues of the ground, and by the appearence of the crop circles, which are younger and have a lighter color in the MSI. In the Ivanpah Playa image, a significant overall difference is observed between the sand colors in both images.

\begin{figure}[thb]
	\centering
	\includegraphics[width=\linewidth,trim={0cm 0cm 0cm 0cm},clip]{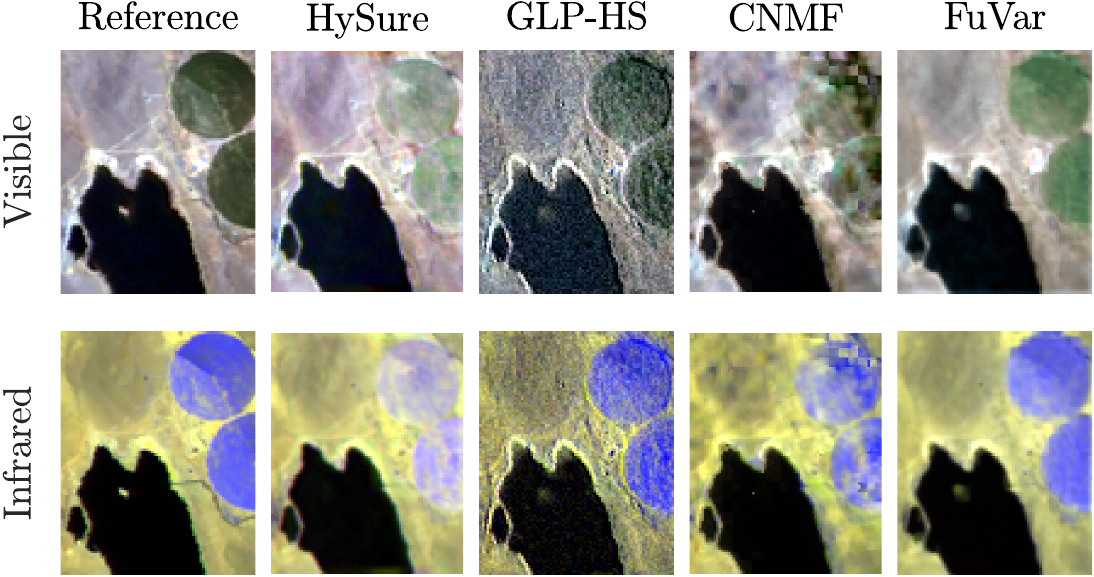}
    \caption{Visual results for the Lake Tahoe hyperspectral image in Example~3.}
    \label{fig:ex3a_hsi_visual_b}
\end{figure}

\begin{figure}[thb]
	\centering
	\includegraphics[width=\linewidth,trim={0cm 0cm 0cm 0cm},clip]{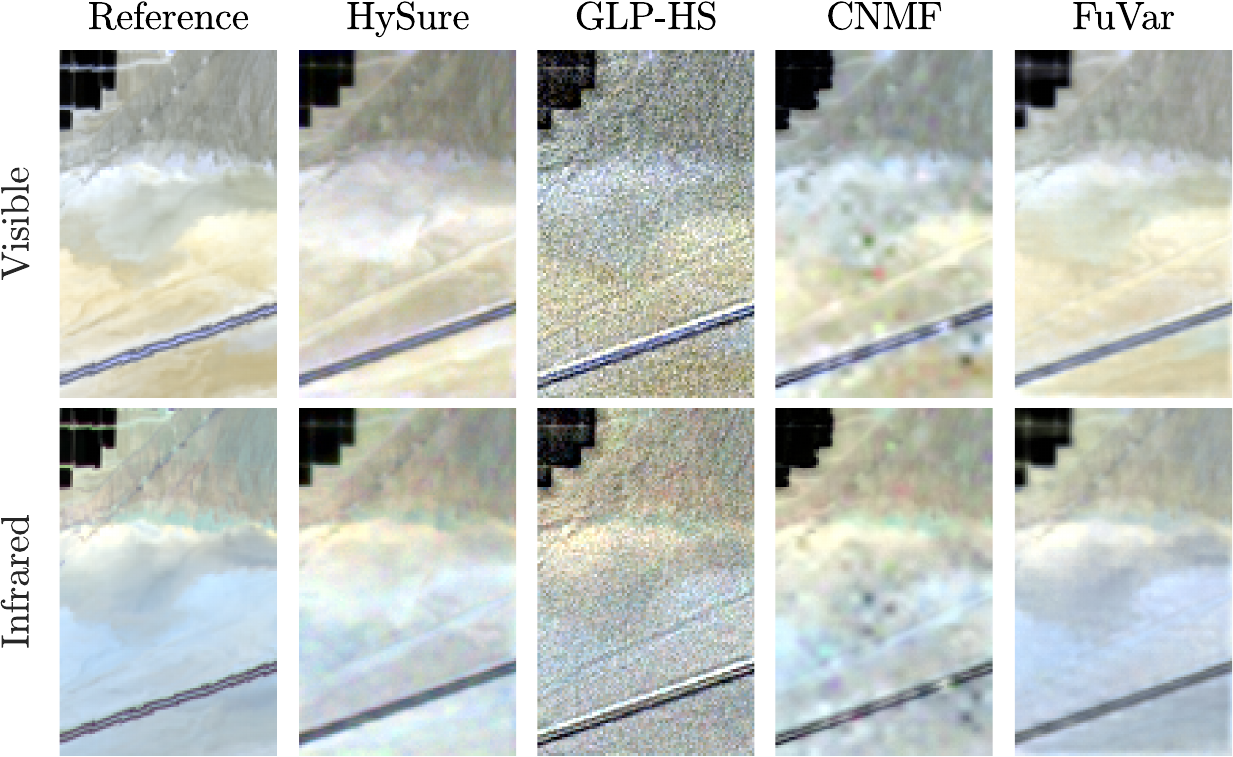}
    \caption{Visual results for the Ivanpah Playa hyperspectral image in Example~3.}
    \label{fig:ex3b_hsi_visual_b}
\end{figure}

The quantitative results for both images are displayed in Table~\ref{tab:ex3_res_b}, and a visual assessment of the results for the super-resolved HS images $\bZ_h$ is presented in Figs.~\ref{fig:ex3a_hsi_visual_b} and~\ref{fig:ex3b_hsi_visual_b}.

\begin{table} [htb]
\caption{Example~3 - Results for Lake Tahoe and Ivanpah Playa images.}
\vspace{-0.15cm}
\centering
\renewcommand{\arraystretch}{1.3}
\setlength\tabcolsep{3.5pt}
\resizebox{\linewidth}{!}
{\begin{tabular}{l|cccc|cccccccccccccccc}
\hline
& \multicolumn{4}{c|}{Lake Tahoe image} & \multicolumn{4}{c}{Ivanpah Playa image} \\
\hline
& PSNR & SAM & ERGAS & UIQI & PSNR & SAM & ERGAS & UIQI \\
\hline
GLP-HS & 18.80 &  9.695 & 6.800 & 0.763 & 19.85 & 3.273 & 3.562 & 0.460 \\
HySure & 17.85 & 11.133 & 6.679 & 0.769 & 22.01 & 2.237 & 2.613 & 0.521 \\
CNMF   & 20.98 &  6.616 & 5.381 & 0.835 & 24.91 & 1.584 & 1.979 & 0.736 \\
FuVar  & \textbf{25.10} &  \textbf{5.530} & \textbf{3.307} & \textbf{0.933} & \textbf{29.70} & \textbf{1.326} & \textbf{1.136} & \textbf{0.922} \\
\hline
\end{tabular}}
\label{tab:ex3_res_b}
\end{table}


\paragraph{Discussion}

The results in Table~\ref{tab:ex3_res_b} show a clear advantage of the FuVar algorithm over the competing methods, with significant performance gains in all metrics for both datasets.
Visual inspection of the recovered super resolved HSIs in Figs.~\ref{fig:ex3a_hsi_visual_b} and~\ref{fig:ex3b_hsi_visual_b} reveals a clear accuracy improvement obtained by using FuVar over the competing algorithms for both the visible and infrared portions of the spectrum. Although the GLP-HS algorithm yielded sharper contours around the circular plantations in Fig.~\ref{fig:ex3a_hsi_visual_b} and the road in Fig.~\ref{fig:ex3b_hsi_visual_b}, it clearly introduced many visual artifacts that do not exist in the reference images. Furthermore, the colors in the GLP-HS results are significantly different from those in the reference image. The FuVar algorithm results show a good compromise between the sharp discontinuities and the smoothness of the more continuous parts of the image.

\subsection{Execution Times}

The execution times for all tested algorithms are shown in Table~\ref{tab:ex4_res}. The higher FuVar execution times are expected since the other fusion methods do not account for spectral variability between the HS and MS images, thus leading to much simpler optimization problems.

\begin{table} [htb]
\caption{Execution times of the algorithms (in seconds).}
\vspace{-0.15cm}
\centering
\renewcommand{\arraystretch}{1.3}
\setlength\tabcolsep{3.5pt}
{\begin{tabular}{l|cccccc}
\hline
& Synthetic 
& Paris 
& \begin{tabular}{@{}c@{}} Lake \\[-0.1cm] Tahoe\end{tabular} 
& \begin{tabular}{@{}c@{}} Ivanpah \\[-0.1cm] Playa\end{tabular} 
& \begin{tabular}{@{}c@{}} Lake \\[-0.1cm] Isabella\end{tabular} 
& Lockwood \\
\hline
GLP-HS & 8.9 & 1.7 & 4.1 & 5.2 & 3.2 & 3.8 \\
HySure & 7.5 & 3.8 & 6.3 & 8.2 & 4.2 & 5.2 \\
CNMF   & 5.1 & 3.8 & 3.9 & 6.5 & 2.5 & 3.1 \\
FuVar  & 132.3 & 155.0 & 391.6 & 248.5 & 263.1 & 262.4 \\
\hline
\end{tabular}}
\label{tab:ex4_res}
\end{table}

\subsection{Parameter Sensitivity}

\begin{figure}[thb]
	\centering
	\includegraphics[width=0.8\linewidth,trim={0cm 0cm 0cm 0cm},clip]{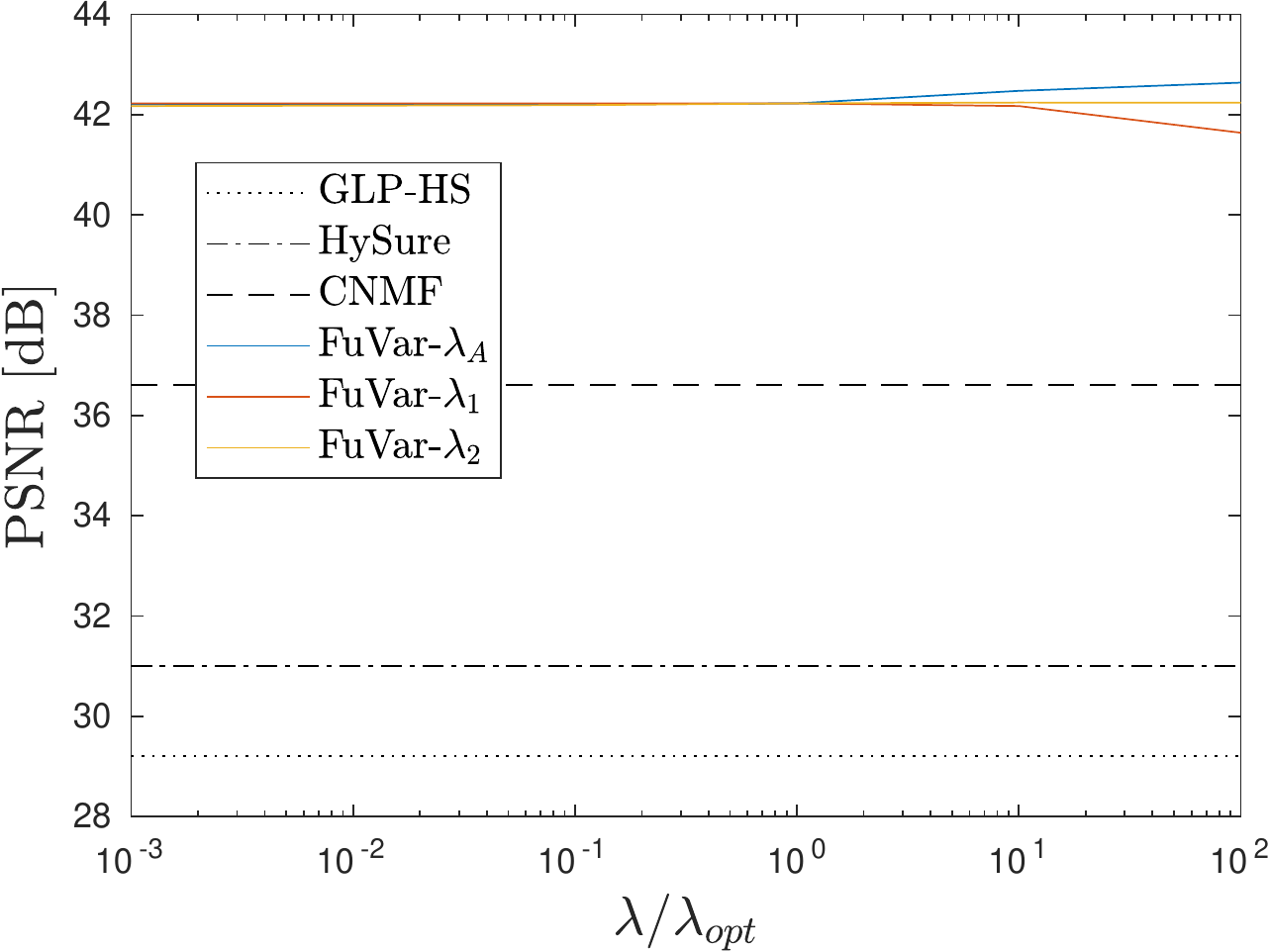}
	\vspace{-0.3cm}
    \caption{Sensitivity of the proposed method with respect to parameters $\lambda_A$, $\lambda_1$, $\lambda_2$ (the horizontal axis denotes the normalized parameter variation).}
    \label{fig:ex_param_sensitivity}
\end{figure}

In this section we explore the sensitivity of the proposed method to the choice of values for parameters $\lambda_A$, $\lambda_1$, $\lambda_2$. For the experimental setup of Example~1, we varied each parameter individually, while keeping the remaining ones fixed at the values described in Section~\ref{sec:exp_setup}, and computed the PSNR of the HSI reconstructed by the FuVar algorithm. The results are shown in Fig.~\ref{fig:ex_param_sensitivity} as a function of the ratio $\lambda/\lambda_{opt}$, where $\lambda_{opt}$ is the empirically selected value of the corresponding parameter. The PSNRs obtained using competing methods are also shown for reference.
It is clear that even variations of parameter values by various orders of magnitude had only a relatively small effect on the resulting PSNR, which remains significantly higher than that of the remaining methods.
This indicates that the performance of the proposed method is not overly sensitive to the choice of the regularization parameters.

\section{Conclusions} \label{sec:conclusions}


We presented a novel hyperspectral and multispectral image fusion strategy accounting for spectral mismatches between the two images, which are often related to illumination and/or seasonal changes. The proposed strategy combines the image fusion paradigm with a parametric model for material spectra, resulting in a new algorithm named FuVar. Simulations with synthetic and real data illustrated the performance improvement obtained by incorporating spectral variability in the fusion problem. The proposed strategy was shown to be quite insensitive to the choice of hyperparameters, and to be competitive with state of the art algorithms even when spectral variability is not present. This allows for broad application of the proposed method.

\appendices

\section{Optimization of the ADMM for the $\bA$ subproblem} \label{sec:appendix1}

In this section, we describe the optimization of the scaled augmented Lagrangian~$\mathcal{L}(\bTheta)$ in~\eqref{eq:admm_lagrangean} with respect to each one of the variables in~${\bTheta}=\{\bw,\bB_1,\bB_2,\bB_3,\bB_4,\bB_5,\bB_6,\bu\}$.

\subsection{Optimizing w.r.t. $\bw$}\label{sec:opt_w}

The subproblem of minimizing $\mathcal{L}(\bTheta)$ with respect to $\bw$ can be extracted from~\eqref{eq:admm_lagrangean} by selecting the terms that depend on~$\bw$:
\begin{align} \label{eq:admm_subp_a_mu_i}
    \min_{\bw} & \,\,\,
    \frac{1}{2}\|\emph{vec}(\bY_{\!m}) 
    - \emph{rep}_M(\bR(\bPsi\circ\bM_h)) \bw\|^2
	\nonumber \\ &
    + \frac{\rho}{2}\Big(
    \|\emph{vec}(\bB_2) - \emph{rep}_M(\bM_h) \bw + \bu_2\|^2 
    \nonumber \\ & 
    + \|\emph{vec}(\bB_3) - \bw + \bu_3\|^2
    \nonumber \\ & 
    + \|\emph{vec}(\bB_6) - \bw + \bu_6\|^2 \Big).
\end{align}

Alternatively, since $\bw$ is the vectorization of the abundance matrix $\bA$, we can re-write this problem for each pixel as
\begin{align} \label{eq:admm_subp_a_mu_ii}
	\min_{\ba_n} & \,\,\, \sum_{n=1}^M \Big(
    \frac{1}{2}\|\by_{m,n} - \bR(\bPsi\circ\bM_h) \ba_n \|^2
    \nonumber \\  &
    + \frac{\rho}{2} \big( 
    \|\bb_{2,n} - \bM_h\ba_n + \bu_{2,n} \|^2
    \nonumber \\ &
    + \|\bb_{3,n} - \ba_n + \bu_{3,n}\|^2
    \nonumber \\ &
    + \|\bb_{6,n} - \ba_n + \bu_{6,n}\|^2
    \big)
    \Big).
\end{align}

Taking the gradients w.r.t. $\ba_n$ and setting them equal to zero leads to
\begin{align} \label{eq:admm_subp_a_mu_iii_b}
        & \Big( \big(\bR(\bPsi\circ\bM_h)\big)^\top 
        \big(\bR(\bPsi\circ\bM_h)\big) 
        + \rho\bM_h^\top\bM_h + 2\rho\bI \Big)
        \ba_n 
        \nonumber \\ & \hspace{0.5cm}
        {}={} \rho\big(
        \bM_h^\top(\bb_{2,n}+\bu_{2,n})
        + (\bb_{3,n}+\bu_{3,n})
        \nonumber \\ & \hspace{0.9cm}
        + (\bb_{6,n} + \bu_{6,n})
        \big)
        + \big(\bR(\bPsi\circ\bM_h)\big)^\top \by_{m,n}
\end{align}
and the solution is given by
%
\begin{align} \label{eq:admm_subp_a_mu_iv}
	\ba^*_n {}={} \bOmega_1^{-1} \bOmega_2
\end{align}
with
\begin{align}
	\bOmega_1 & {}={} \big(\bR(\bPsi\circ\bM_h)\big)^\top         \big(\bR(\bPsi\circ\bM_h)\big) + \rho\bM_h^\top\bM_h + 2\rho\bI 
    \nonumber \\
    \bOmega_2 & {}={} \rho\big(
        \bM_h^\top(\bb_{2,n}+\bu_{2,n})
        + (\bb_{3,n}+\bu_{3,n})
        + (\bb_{6,n} + \bu_{6,n})
        \big)
        \nonumber \\ & \hspace{0.5cm}
        + \big(\bR(\bPsi\circ\bM_h)\big)^\top \by_{m,n}\,.
\end{align}

\subsection{Optimizing w.r.t. $\bB_1$}

Equivalently to Section~\ref{sec:opt_w}, the subproblem for $\bB_1$ can be written as
\begin{align} \label{eq:admm_subp_b1_i}
	\min_{\bB_1} & \,\,\,
	\frac{1}{2}\|\bY_{\!h} - \bB_1\|_F^2
    \nonumber \\ & 
	+ \frac{\rho}{2}
    \|\emph{rep}_L(\bD^\top\bF^\top)\emph{vec}(\bB_2^\top)
    - \emph{vec}(\bB_1^\top) + \bu_1 \|^2 .
\end{align}

Representing the last term of~\eqref{eq:admm_subp_b1_i} in matrix form leads to
\begin{align} \label{eq:admm_subp_b1_ii}
	\min_{\bB_1} & \,\,\,
	\frac{1}{2}\|\bY_{\!h} - \bB_1\|_F^2
    \nonumber \\ & 
	+ \frac{\rho}{2}
    \|\bD^\top\bF^\top\bB_2^\top - \bB_1^\top 
    + \emph{vec}^{-1}(\bu_1) \|^2_F
\end{align}
where $\big(\emph{vec}^{-1}(\bu_1)\big)^\top=\widetilde{\bU}_1$. Problem~\eqref{eq:admm_subp_b1_ii} is equivalent to
\begin{align} \label{eq:admm_subp_b1_iii}
	\min_{\bB_1} & \,\,\, 
	\frac{1}{2}\|\bY_{\!h} - \bB_1\|_F^2
	+ \frac{\rho}{2}
    \|\bB_2\bF\bD - \bB_1 + \widetilde{\bU}_1 \|^2_F
\end{align}
and can be re-written individually for each pixel as
\begin{align} \label{eq:admm_subp_b1_iv}
	\min_{\bB_1} & \,\,\, \sum_{n=1}^N \Big(
	\frac{1}{2}\|\by_{h,n} - \bb_{1,n}\|^2
    \nonumber \\ & 
	+ \frac{\rho}{2}
    \|[\bB_2\bF\bD]_n - \bb_{1,n} + \widetilde{\bu}_{1,n}\|^2
    \Big)\,.
\end{align}

Thus, taking the derivative of~\eqref{eq:admm_subp_b1_iv} for each pixel and setting it equal to zero, we have
\begin{align} \label{eq:admm_subp_b1_v}
	\bb_{1,n}^\ast = \frac{1}{1+\rho} \big(
    \by_{y,n} + \rho[\bB_2\bF\bD]_n 
    + \rho \widetilde{\bu}_{1,n}
    \big)
\end{align}
for $n=1,\ldots,N$.

\subsection{Optimizing w.r.t. $\bB_2$}

Analogously to Section~\ref{sec:opt_w}, the subproblem for $\bB_2$ can be written as
\begin{align} \label{eq:admm_subp_Zh_i}
	\min_{\bB_2} &
    \frac{\rho}{2}\Big(
    \|\emph{rep}_L(\bD^\top\bF^\top)\emph{vec}(\bB_2^\top) 
    - \emph{vec}(\bB_1^\top) + \bu_1 \|^2
    \nonumber \\ & 
    + \|\emph{vec}(\bB_2) - \emph{rep}_M(\bM_h) \bw + \bu_2\|^2 
    \Big)\,.
\end{align}

Representing the variables in matrix form, problem \eqref{eq:admm_subp_Zh_i} is equivalent to
\begin{align} \label{eq:admm_subp_Zh_ii}
	\min_{\bB_2}  &  \,\,\, 
	\frac{\rho}{2}\Big(
    \|\bD^\top\bF^\top\bB_2^\top 
    - \bB_1^\top + \bU_1 \|^2_F
    \nonumber \\ & \hspace{1.5cm}
    + \|\bB_2^\top - \bA^\top\bM_h^\top + \bU_2^\top\|^2_F
    \Big)
\end{align}
%
and can be recast to each HS band as
\begin{align} \label{eq:admm_subp_Zh_iii}
	\min_{\bB_2}  &  \,\,\, \sum_{\ell=1}^L
    \frac{1}{2} \Big(
    \| \, \bD^\top\bF^\top[\bB_2^\top]_\ell  
    - [\bB_1^\top]_\ell
    + [\bU_1]_{\ell} \|^2
    \nonumber \\ & \hspace{1.5cm} 
    \|\, [\bB_2^\top]_\ell - [\bA^\top\bM_h^\top]_\ell
    + [\bU_2^\top]_\ell \|^2
    \Big).
\end{align}

Taking the derivative w.r.t. each band and setting it equal to zero gives
\begin{align} \label{eq:admm_subp_Zh_iv}
	\Big( \bI + \bF\bD \bD^\top\bF^\top 
    \Big) [\bB_2^\top]_\ell 
	{}={} &
    [\bA^\top\bM_h^\top]_\ell - [\bU_2^\top]_\ell
    + \bF\bD [\bB_1^\top]_\ell
    \nonumber \\ & %
    - \bF\bD [\bU_1]_{\ell}
\end{align}
for $\ell=1,\ldots,L$. Since matrices $\bF$ and $\bD$ are sparse, problem~\eqref{eq:admm_subp_Zh_iv} can be solved efficiently using iterative methods such as the Conjugate Gradient method, where the large matrices involved can be computed implicitly.


\subsection{Optimizing w.r.t. $\bB_3$}

This subproblem is equivalent to
\begin{align} \label{eq:admm_subp_b3_i}
	& \min_{\bB_3}  \,\,\,
	\frac{\rho}{2}\Big(
    \|\emph{vec}(\bB_3) - \bw + \bu_3\|^2
    \nonumber \\ & \hspace{1.5cm}
    + \|\emph{vec}(\bB_4) - \bH_h \emph{vec}(\bB_3) + \bu_4\|^2
    \nonumber \\ & \hspace{1.5cm}
    + \|\emph{vec}(\bB_5) - \bH_v \emph{vec}(\bB_3) + \bu_5\|^2
    \Big)
\end{align}
%
whose solution w.r.t. $\emph{vec}(\bB_3)$ leads to
\begin{align} \label{eq:admm_subp_b3_ii}
	\emph{vec}(\bB_3)^\ast {}={} & \big( \bI 
    + \bH_h^\top\bH_h + \bH_v^\top\bH_v\big)^{-1}
    \big(\bw - \bu_3 
    \nonumber \\ & \hspace{1cm}
    + \bH_h^\top\emph{vec}(\bB_4) + \bH_h^\top\bu_4
    \nonumber \\ & \hspace{1cm}
    + \bH_v^\top \emph{vec}(\bB_5) + \bH_v^\top\bu_5
    \big)\,.
\end{align}

The dimensionality of the matrices $\bH_h$ and $\bH_v$ in this problem makes the direct inversion of the terms involving them in \eqref{eq:admm_subp_b3_ii} impractical.
However, assuming these differential operators are block circulant, and following the same approach as in~\cite{drumetz2016blindUnmixingELMMvariability} and~\cite{imbiriba2018glmm}, these operations can be computed efficiently.
This is done by performing the computations with the differential operators and their adjoint in the Fourier domain, independently for each endmember.
Denoting the convolution masks associated with $\bH_h$ and $\bH_v$ by $\bh_h$ and $\bh_v$, respectively (see details in~\cite{drumetz2016blindUnmixingELMMvariability}), assuming circular boundary conditions and using properties of the Fourier transform, the solution in~\eqref{eq:admm_subp_b3_ii} can be computed equivalently as
\begin{align} \label{eq:admm_subp_b3_iii}
	[\bbB_3]_{:,:,p} {}={} \calF^{-1} & \Big(\big(
    \calF(\bbA_{:,:,p} - [\bbU_3]_{:,:,p})
    \nonumber \\ &
    + \calF^\ast(\bh_h) 
    \circ \calF([\bbB_4]_{:,:,p}+[\bbU_4]_{:,:,p})
    \nonumber \\ &
    + \calF^\ast(\bh_v)
    \circ \calF([\bbB_5]_{:,:,p}+[\bbU_5]_{:,:,p})
    \big) 
    \nonumber \\ &
    \oslash \big( \cb{1}_{M_1} \cb{1}_{M_2}^\top 
    + |\calF(\bh_h)|^2 + |\calF(\bh_v)|^2 \big) \Big)
\end{align}
for $p=1,\ldots,P$, where $\calF$ and $\calF^{-1}$ are the forward and inverse 2D Fourier transforms, $\oslash$ is the elementwise division, and tensors $\bbA$, $\bbU_3$, $\bbB_4$, $\bbU_4$, $\bbB_5$, $\bbU_5$, of size $M_1\times M_2 \times P$ are the cube-ordered versions of the respective variables in~\eqref{eq:admm_subp_b3_ii}.
$\calF^\ast$ is the complex conjugate of the Fourier transform, and $|\cdot|$ is the modulus operator, applied elementwise to a matrix.


\subsection{Optimizing w.r.t. $\bB_4$}

The equivalent optimization problem can be written as
\begin{align} \label{eq:admm_subp_b4_i}
	\min_{\bB_4}  & \,\,\, 
	\lambda_A \|\bB_4\|_{2,1}
	+ \frac{\rho}{2}
    \|\emph{vec}(\bB_4) - \bH_h \emph{vec}(\bB_3) - \bu_4\|^2\!.
\end{align}

Following the same idea as in~\cite{drumetz2016blindUnmixingELMMvariability}, the solution of problem \eqref{eq:admm_subp_b4_i} is equivalent to the proximal operator of the $L_2$ norm. The solution to this problem is obtained by using block soft thresholding, which can then be expressed as
\begin{align} \label{eq:admm_subp_b4_ii}
	\bB_4^\ast {}={} \emph{soft}_{\lambda_A/\rho}
    \big( \bH_h\emph{vec}(\bB_3) + \bu_4 \big)
\end{align}
where the $\emph{soft}$ operator for the $L_2$ norm is defined as~\cite{drumetz2016blindUnmixingELMMvariability}
\begin{align}
	\emph{soft}_a(\bb) {}={} 
    \max\Big(1-\frac{a}{\|\bb\|}, \, 0\Big) \,\bb
\end{align}
with $\emph{soft}_a(\cb{0})=\cb{0}$, $\forall a$.

\subsection{Optimizing w.r.t. $\bB_5$}

Equivalently, this optimization problem can be written as
\begin{align} \label{eq:admm_subp_b5_i}
	\min_{\bB_5}  & \,\,\, 
	\lambda_A \|\bB_5\|_{2,1}
	+ \frac{\rho}{2}
    \|\emph{vec}(\bB_5) - \bH_v \emph{vec}(\bB_3) + \bu_5\|^2
\end{align}
which is identical to problem~\eqref{eq:admm_subp_b4_ii}. Following the strategy outlined above, the solution to \eqref{eq:admm_subp_b5_i} is given by
\begin{align} \label{eq:admm_subp_b5_ii}
	\bB_5^\ast {}={} \emph{soft}_{\lambda_A/\rho}
    \big( \bH_v\emph{vec}(\bB_3) - \bu_5 \big).
\end{align}

\subsection{Optimizing w.r.t. $\bB_6$}

This optimization problem can be written as
\begin{align} \label{eq:admm_subp_b6_i}
	\min_{\bB_6}  & \,\,\, 
    \iota_+(\bB_6) 
	+ \frac{\rho}{2} \|\emph{vec}(\bB_6) - \bw + \bu_6\|^2.
\end{align}

The solution to this problem is presented in~\cite{drumetz2016blindUnmixingELMMvariability}, and is given by
\begin{align} \label{eq:admm_subp_b6_ii}
	\bB_6^\ast {}={} \max\Big(
    \emph{vec}^{-1}(\bw) - \emph{vec}^{-1}(\bu_6)
    , \, \cb{0}\Big).
\end{align}

\subsection{Dual update $\bu$}

The dual update is given by
\begin{align}
    \bu^{(k+1)} & {}={} \bu^{(k)} + \mathcal{A}\bx^{(k+1)} 
    + \mathcal{B}\bw^{(k+1)} - \bc
\end{align}
which becomes
\begin{align}
\begin{split}
	\bu_1 & {}\leftarrow{} \bu_1 - \emph{vec}(\bB_1^\top) + \emph{rep}_L(\bD^\top\bF^\top)\emph{vec}(\bB_2^\top)
    \\ 
    \bu_2 & {}\leftarrow{} \bu_2 + \bP_{\top}\emph{vec}(\bB_2^\top) - \emph{rep}_M(\bM_h)\bw
    \\ 
    \bu_3 & {}\leftarrow{} \bu_3 + \emph{vec}(\bB_3) - \bw
    \\ 
    \bu_4 & {}\leftarrow{} \bu_4 + \emph{vec}(\bB_4) - \bH_h\emph{vec}(\bB_3)
    \\ 
    \bu_5 & {}\leftarrow{} \bu_5 + \emph{vec}(\bB_5) - \bH_v\emph{vec}(\bB_3)
    \\ 
    \bu_6 & {}\leftarrow{} \bu_6 + \emph{vec}(\bB_6) - \bw\,.
\end{split}
\end{align}







\section{Optimization of the ADMM for the $\bPsi$ subproblem} \label{sec:appendix2}

In this section, we describe the optimization of the scaled augmented Lagrangian~$\mathcal{L}(\bB,\bPsi,\bU)$ in~\eqref{eq:opt_bpsi_iii} with respect to each one of the variables~$\bB$, $\bPsi$ and $\bU$.

\subsection{Optimizing w.r.t. $\bB$}

The resulting optimization problem for $\bB$ can be written as
\begin{align} \label{eq:opt_bpsi_subb_i}
	\mathop{\min}_{\bB} \,\, &
    \frac{1}{2}\|\bY_{\!m} - \bR\bB\bA\|_F^2
    + \frac{\rho}{2} \| \bB - \bPsi\circ\bM_h + \bU\|_F^2 .
\end{align}

Taking the derivative with respect to $\bB$ and setting it equal to zero leads to
\begin{align} \label{eq:opt_bpsi_subb_ii}
	\frac{1}{\rho} \bR^\top\bR \bB \bA\bA^\top 
    \!\! + \bB 
    {}={} & 
    \frac{1}{\rho} \bR^\top\bY_{\!m} \bA^\top
    \!\! + \bPsi \circ \bM_h
    - \bU.
\end{align}
This consists in the Sylvester equation $AXB-X+C=0$.\footnote{In MATLAB$^{\text{TM}}$, X = dlyap(A,B,C) solves the  equation AXB - X + C = 0, with A, B, and C with compatible dimensions but not required to be square.}

\subsection{Optimizing w.r.t. $\bPsi$}

The optimization problem can be written as
\begin{align} \label{eq:opt_bpsi_subp_i}
	\min_{\bPsi\geq0}   \,\,\, &    \frac{\lambda_1}{2} 
    \|\bPsi - \cb{1}_L\cb{1}_P^\top \|_F^2
    + \frac{\lambda_2}{2} \|\bH_{\!\ell}\bPsi\|_F^2
    \\ & \nonumber \quad
    + \frac{\rho}{2} \| \bB - \bPsi\circ\bM_h + \bU\|_F^2\,.
\end{align}

For simplicity, the nonnegativity constraint will be initially ignored. Taking the derivative w.r.t. $\bPsi$ and setting it equal to zero leads to:
\begin{align} \label{eq:opt_bpsi_subp_ii}
	& 
    \lambda_1(\bPsi-\cb{1}_L\cb{1}_P^\top)
    + \lambda_2\bH_{\!\ell}^\top\bH_{\!\ell} \bPsi
    \nonumber \\ &
    - \rho (\bM_h\circ\bB + \bM_h\circ\bU - \bM_h\circ\bM_h\circ\bPsi)
    = \cb{0}\,.
\end{align}

The matrix equation~\eqref{eq:opt_bpsi_subp_ii} can be written in vector form as
\begin{align} \label{eq:opt_bpsi_subp_iii}
	& 
    \Big(\lambda_1\bI  
    + \lambda_2 \emph{rep}_P(\bH_{\!\ell}^\top\bH_{\!\ell})
    + \rho \,\emph{diag}(\emph{vec}(\bM_h\circ\bM_h)) 
    \Big) 
    \nonumber \\ & \hspace{5.5ex}
    \times \emph{vec}(\bPsi)  = \lambda_1\cb{1}
    + \rho\, \emph{vec}(\bM_h\circ\bB+\bM_h\circ\bU)
\end{align}
which is a system of equations that can be solved efficiently since the matrices on the left-hand side are very sparse. Finally, positivity is introduced in the solution obtained by solving the sparse system in~\eqref{eq:opt_bpsi_subp_iii} as
\begin{align} \label{eq:opt_bpsi_subp_iv}
	\bPsi^* {}={} \max(\bPsi,\cb{0}) \,.
\end{align}

\vspace{-0.8cm}
\subsection{Dual update $\bu$}

The dual update is given by
\begin{align}
	\bU \leftarrow \bU + \bB - \bM_h \circ \bPsi\,.
\end{align}






\bibliographystyle{IEEEtran}
\bibliography{references,references_fusion}

\vspace{-1cm}
\begin{IEEEbiographynophoto}{Ricardo Augusto Borsoi (S'18)} 
received the MSc degree in electrical engineering from Federal University of Santa Catarina (UFSC), Florian\'opolis, Brazil, in 2016. He is currently working towards his doctoral degree at Universit\'e C\^ote d'Azur (OCA) and at UFSC. His research interests include image processing, tensor decomposition, and hyperspectral image analysis.
\end{IEEEbiographynophoto}

\vspace{-1cm}
\begin{IEEEbiographynophoto}{Tales Imbiriba (S'14, M'17)}   
received his Doctorate degree from the Department of Electrical Engineering (DEE) of the Federal University of Santa Catarina (UFSC), Florian\'opolis, Brazil, in 2016. He served as a Postdoctoral Researcher at the DEE--UFSC and is currently a Postdoctoral Researcher at the ECE dept. of the Northeastern University, Boston, MA, USA. 
His research interests include audio and image processing, pattern recognition, kernel methods, adaptive filtering, and Bayesian Inference.
\end{IEEEbiographynophoto}

\vspace{-1cm}
\begin{IEEEbiographynophoto}{Jos\'e Carlos M. Bermudez (S'78,M'85,SM'02)}
received the Ph.D. degree from Concordia University, Canada, in 1985. He is a Professor at the Department of Electrical Engineering, Federal University of Santa Catarina, Brazil, and at the Graduate Program on Electronic Engineering and Computing, Catholic University of Pelotas, Brazil. His research interests are in statistical signal processing, including image processing, adaptive filters, hyperspectral image processing and machine learning. He is Senior Area Editor of the IEEE Transactions on Signal Processing and chair of the Signal Processing Theory and Methods technical committee for the IEEE Signal Processing Society. 

\end{IEEEbiographynophoto}

\end{document}